\documentclass[10pt]{article} 

\usepackage[preprint]{tmlr}

\usepackage{amsmath,amsfonts,bm}
\usepackage{mathtools}
\usepackage{amsthm}
\usepackage{tikz}
\usetikzlibrary{arrows.meta, positioning, fit}

\theoremstyle{definition}
\newtheorem{definition}{Definition}
\theoremstyle{definition}
\newtheorem{example}{Example}[section]

\newcommand{\captiona}{{\em (a)}}
\newcommand{\captionb}{{\em (b)}}

\newcommand{\independent}{\mathrel{\perp\!\!\!\perp}}

\def\eqref#1{equation~\ref{#1}}

\def\1{\bm{1}}

\DeclareMathAlphabet{\mathsfit}{\encodingdefault}{\sfdefault}{m}{sl}
\SetMathAlphabet{\mathsfit}{bold}{\encodingdefault}{\sfdefault}{bx}{n}

\newcommand{\R}{\mathbb{R}}

\usepackage{inconsolata}
\usepackage{hyperref}
\usepackage{url}
\usepackage{algorithm}
\usepackage{algpseudocode}
\usepackage{caption}
\usepackage{wrapfig}
\usepackage{tcolorbox}
\usepackage{booktabs}
\usepackage[frozencache, cachedir=minted-cache]{minted}
\usepackage{cleveref}

\title{Causal Foundation Models}

\author{\name Christopher Stith\thanks{Equal contribution.} \email christopher@layer6.ai \\
      \addr Layer 6 AI, Toronto, Canada
      \AND
      \name Hossein Rahmani\footnotemark[1] \email hossein.rahmani@td.com \\
      \addr TD Bank Group, Toronto, Canada
      \AND
      \name Jesse C. Cresswell \email jesse@layer6.ai\\
      \addr Layer 6 AI, Toronto, Canada
      }

\newcommand{\jupyter}[1]{\href{#1}{\begingroup
\setbox0=\hbox{\includegraphics[height=1.25em]{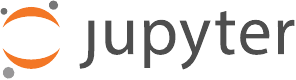}}%
\parbox{\wd0}{\box0}\endgroup}}

\begin{document}

\maketitle

\begin{abstract}
Causal inference is the practice of estimating the effect of a treatment or intervention from data. It traditionally requires a bespoke pipeline for every new problem: first proposing a causal mechanism, selecting a compatible estimator, and finally training it. Meanwhile, across diverse settings and modalities, much of machine learning has shifted to the paradigm of foundation models: networks pretrained once at scale and applied to new tasks without fine-tuning. \emph{Causal foundation models} (CFMs) bring this paradigm to causal inference. CFMs are pretrained neural networks that estimate causal quantities, such as the average treatment effect, on entirely new datasets using in-context learning without requiring model updates. This work provides a practical introduction to this emerging area. We summarize the necessary background in causal inference and machine learning before discussing CFMs. Throughout, we include example code and Jupyter notebooks, which can be accessed by clicking on the \jupyter{https://github.com/layer6ai-labs/cfms/blob/main/notebooks/Foundation_models_quickstart.ipynb} icons. The full codebase is available at \href{https://github.com/layer6ai-labs/cfms}{\texttt{github.com/layer6ai-labs/cfms}}.
\end{abstract}

\section{Introduction}\label{sec:intro}
Causal inference is the practice of estimating causal effects between variables \citep{pearl2009causalinferenceinstatisticsanoverview, imbens2015causalinference}. For example: How effective is a medication at preventing a given disease? Did a policy cause a decrease in unemployment? If a central bank raises interest rates, what will the effect be on consumer spending? Causal inference attempts to answer these questions while avoiding the common pitfall of mistaking association with causation, captured in the adage that ``correlation is not causation''. Causal inference has applications across many domains, such as economics and policy-making \citep{athey2017thestateofappliedeconometrics, chernozhukov2018double}, marketing \citep{bottou2013counterfactualreasoningandlearning, gordon2019comparisonofapproaches}, and medicine \citep{alaa2017bayesianinferenceofindividualized, shalit2017estimatingindividualtreatmenteffects}.

The causal inference community has built a rich library of estimation methods, including Bayesian additive regression trees \citep{chipman2010bart}, double machine learning \citep{chernozhukov2018double}, causal forests \citep{wager2018}, as well as S-, T-, and X-learners \citep{kunzel2019metalearners}. To approach any given problem with most of these methods, one must first study the data and propose an underlying causal mechanism, choose an estimator that fits this mechanism, tune hyperparameters on validation data, and only then train the final estimator on the data. For each new problem, the full pipeline must be repeated, with no opportunity to reuse tuned models or transfer knowledge between tasks. 

\begin{figure}[t]
    \centering
    \begin{minipage}{0.60\textwidth}
        \includegraphics[width=\textwidth]{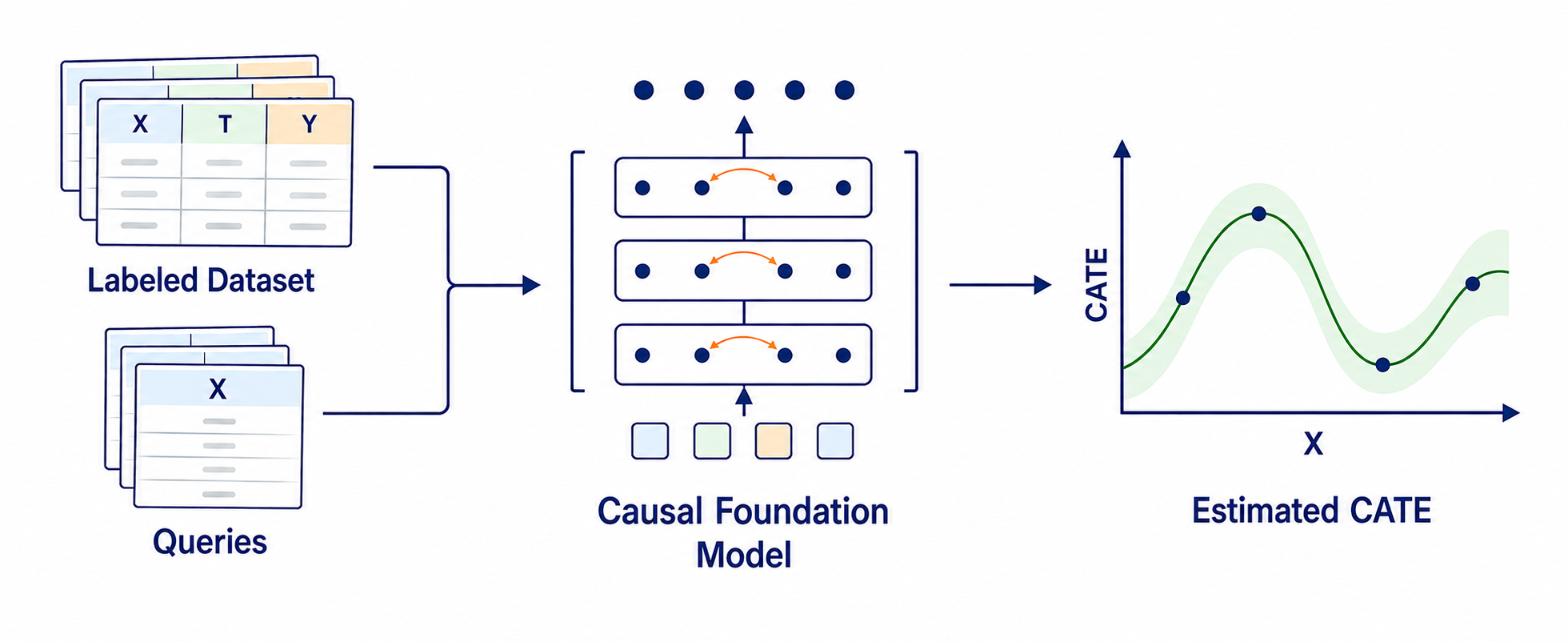}
    \end{minipage}
    \hfill
    \begin{minipage}{0.39\textwidth}
        \begin{minted}[fontsize=\footnotesize, bgcolor=white]{python}
model = CATEEstimator()
# Calling .fit() only loads data into context; 
# no training required
model.fit(X_ctx, T_ctx, Y_ctx)
cate = model.estimate_cate(X_qry)
        \end{minted}
    \end{minipage}
    \captionsetup{width=1\textwidth}
    \caption{Causal foundation models predict on each new dataset using in-context learning without training or fine-tuning. Labeled context data (\texttt{X\_ctx}, etc.) are provided as examples so that the model can predict causal effects on unlabeled query data (\texttt{X\_qry}). A quickstart example is available in this notebook: \jupyter{https://github.com/layer6ai-labs/cfms/blob/main/notebooks/Foundation_models_quickstart.ipynb}.
    }
    \label{fig:code-demo}
\end{figure}

Recently, \emph{causal foundation models} have emerged as a strong and efficient alternative. These are pretrained neural networks that can be applied immediately to any causal inference task \emph{without further training or fine-tuning}. The heart of causal foundation models (CFMs) lies at the intersection of causal inference and \emph{learning-to-learn}, in which models learn the ability to predict causal effects in unseen settings from observational data.\footnote{In machine learning, learning-to-learn is often called \emph{meta-learning} \citep{finn2017modelagnosticmetalearning}. However, meta-learning has its own meaning in causal inference \citep{kunzel2019metalearners, frauen2026machinelearningforcausalinference}, so we avoid using this term in this work.} They are trained on tasks sampled from a prior over possible data-generating processes and causal mechanisms, learning to estimate causal effects in a wide variety of scenarios. At inference time, they leverage in-context learning on labeled examples to perform amortized Bayesian inference and make causal estimates (\Cref{fig:code-demo}). 

In this work, we provide a practical and hands-on introduction to CFMs. Our main goal is to provide the reader with the information, tools, and examples they need to use CFMs in their own work. We include example code and Jupyter notebooks throughout this paper to help the reader start using these models immediately. These resources can be accessed by clicking on the \jupyter{https://github.com/layer6ai-labs/cfms/tree/main} icons throughout. We also release our full codebase at \href{https://github.com/layer6ai-labs/cfms/tree/main}{\texttt{github.com/layer6ai-labs/cfms}}.

CFMs have demonstrated top performance on causal inference tasks, and we expect that time will only demonstrate further transformational applications. What is remarkable is that these models bring not only a vast increase in inference speed, but also improved performance. This work exists to introduce CFMs to a wider audience, compare the various approaches that have been taken for CFM design, discuss the latest developments in the area, and invite researchers and practitioners to give them a try.

We begin by discussing the necessary background material from causal inference and machine learning in \Cref{sec:background} before jumping into the core material on CFMs in \Cref{sec:cfms-core}. In \Cref{sec:benchmarking}, we benchmark the first three openly available CFMs \citep{robertson2025do-pfn, balazadeh2026causalpfn, ma2026causalfm} together with popular traditional causal inference models. In \Cref{sec:cfms-broad}, we review the many new developments in the field which form promising areas for future growth, and applications of CFMs in other scientific areas.

\section{Background}\label{sec:background}
\subsection{Why causal inference?}\label{subsec:prediction_vs_causal_inference}

When predictive models deliver excellent results, why do we need causal machinery? \citet{mooij2016distinguishingcausefromeffect} give a classic illustration demonstrating the difference between \emph{predictive} and \emph{causal} inference. Across European countries, regions with more storks also tend to have higher human birth rates \citep{matthews2000storks}. Consider a model trained purely to predict birth rate from stork population: it will exploit the correlation and predict well across the observed regions. What might this model predict if we ask it ``What would happen to the birth rate in a region if we doubled the stork population?'' This is an \emph{interventional} question, not merely a predictive one, for which correlational analysis is insufficient. Failing to distinguish these subtleties would lead to a public policy disaster: the predictive model might recommend that in order to increase the human birth rate, a country should import more storks!

In more detail, the predictive model has learned the \emph{observational} relationship between stork population and birth rate, but the policy application requires knowledge of the interventional, or \emph{causal}, relationship. A predictive model answers the question ``What outcome do we expect for this individual, as-is?'' A causal model answers the question ``What outcome would we expect for this individual if we intervened?'' The latter question requires reasoning about the mechanism generating the data, not just observing the data itself. Depending on the underlying mechanism, these two questions can have different or identical answers. Without additional assumptions, there is no way to tell from observational data alone which is the case. In the following discussions, we formalize how one can answer this question.

\subsection{The potential outcomes framework for causal inference}\label{subsec:the-potential-outcomes-framework}

We work in the Neyman-Rubin potential outcomes framework \citep{rubin2005causalinference, imbens2015causalinference}. We let capital letters denote random variables, and lower-case letters denote values they can take. We let $\mathcal{T} \subset \R$ denote the \textit{treatment space} of interventions that can be applied. There are three common options for $\mathcal{T}$:
\begin{itemize}
    \item \textbf{(Binary treatment)} $\mathcal{T} = \{0, 1\}$, e.g. control group vs. treatment group.
    \item \textbf{(Multi-armed treatment)} $\mathcal{T} = \{1, \ldots, K\}$, e.g. a choice among $K$ different medical treatments.
    \item \textbf{(Continuous treatment)} $\mathcal{T}$ is an interval, often normalized to $[0, 1]$. In a financial setting, this could denote a price or interest rate offered to a customer.
\end{itemize}
We let $\mathcal{X}$ denote the covariate space and $\mathcal{Y}$ the outcome space (often either $\R$ or a categorical space). For example, $\mathcal{X} = \mathbb{N} \times [0, \infty)$ could denote the covariate space of $\text{Age} \times \text{Income}$, and $\mathcal{Y} = \{0, 1\}$ could denote whether or not the customer bought a product.

The observed values $X \in \mathcal{X}$, $T \in \mathcal{T}$, and $Y \in \mathcal{Y}$ are random variables. If we are observing a population of $N$ individuals/units, we use subscripts like $X_n, T_n$, and $Y_n$ to denote individual-level variables, i.e. the observed covariates, treatment, and outcome of the $n$th individual. 

The key component of the potential outcomes framework is to assume, for every treatment value $t \in \mathcal{T}$, the existence of a random variable $Y(t)$, which is the outcome which \emph{would occur} under treatment $T = t$. The variables $Y(t)$ are referred to as \emph{potential outcomes}. On an individual level, $Y_n(t)$ denotes the potential outcome for individual $n$ if they were to receive treatment $t$. 

We assume that for all $t \in \mathcal{T}$, 
\begin{equation}\label{eq:consistency_assumption}
    T = t \implies Y = Y(t).
\end{equation}
This is known as the \emph{consistency assumption} \citep{neal2020introduction}. It states that the potential outcomes must agree with what we actually observe. One can imagine the existence of an omniscient oracle that knows every potential outcome for every individual; the consistency assumption states that the oracle's knowledge agrees with whatever happens in the real world when a given treatment is applied.

Unfortunately, we are not omniscient oracles. We can never know all potential outcomes for a given individual. Once individual $n$ is given treatment $t$, we can never know what would have happened if $t' \neq t$ had been administered. This is known as the fundamental problem of causal inference \citep{holland1986statisticsandcausalinference}.

\begin{tcolorbox}[colback=teal!5, colframe=teal!60!black, title=Fundamental Problem of Causal Inference]
\hypertarget{fundamental_problem_of_causal_inference}{}
If an individual's observed treatment and outcome are $t$ and $y$, respectively, then it is impossible to know $Y(t')$ for $t' \neq t$.
\end{tcolorbox}

The observed treatment $T = t$ and outcome $Y = Y(t)$ are called the \emph{factual} treatment and outcome. Any hypothetical alternatives like $t'$ and $Y(t')$ are called \emph{counterfactual} treatments and outcomes. Note that even with complete knowledge of every covariate, $Y(t)$ need not be deterministic; exogenous noise means potential outcomes can vary across individuals that share identical covariates $X$. This is why causal quantities of interest are typically expectations or distributions rather than pointwise counterfactuals.

Instead of always working directly with potential outcomes, different applications call for different summary statistics. Hence, there are several important \emph{causal estimands} we consider:\footnote{We return to a general definition of causal estimands in \Cref{subsec:data_generating_processes_and_identifiability}.}
\begin{itemize}
    \item The \textbf{conditional expected potential outcome} (CEPO) over a population of individuals that share covariates $x$, and are given the same treatment $t$, is 
    \begin{equation}\label{eq:cepo_def}
        \mu_t(x) \coloneqq \mathbb{E}[Y(t) \mid X = x].
    \end{equation}
    The CEPO $\mu_t(x)$ is thus the expected outcome of administering treatment $t$ to an individual with covariates $X = x$. CEPOs are useful for comparing between different treatments, which brings us to the next item.
    \item In the binary treatment setting, the \textbf{average treatment effect} (ATE) is 
    \begin{equation}\label{eq:ate_def}
        \text{ATE} \coloneqq \mathbb{E}[Y(1) - Y(0)],
    \end{equation}
    where the expectation is over $X$. The ATE tells us how much the outcome would change on average were the treatment $t=1$ applied to the entire population, compared to the baseline $t=0$. Notice that the ATE can be expressed in terms of CEPOs, using linearity of expectations, and the law of iterated expectations:
    \begin{align}\label{eq:ate_cepo}
    \begin{aligned}
        \text{ATE} &= \mathbb{E}[Y(1)] - \mathbb{E}[Y(0)] \\
        &= \mathbb{E}[\mathbb{E}[Y(1) \mid X = x]] - \mathbb{E}[\mathbb{E}[Y(0) \mid X = x]] \\
        &= \mathbb{E}[\mu_1(x)] - \mathbb{E}[\mu_0(x)] = \mathbb{E}[\mu_1(x) - \mu_0(x)].
    \end{aligned}
    \end{align}
    
    \item Only looking at the ATE can hide strong effects if some parts of the population are affected differently than others. Hence, we also consider the \textbf{conditional average treatment effect} (CATE) over a population of individuals that share covariates $x$,
    \begin{equation}\label{eq:cate_def}
        \text{CATE}(x) \coloneqq \mathbb{E}[Y(1) - Y(0) \mid X = x] = \mu_1(x) - \mu_0(x).
    \end{equation}
    From this definition, we have $\text{ATE} = \mathbb{E}[\text{CATE}(x)]$. The CATE tells us how large of a causal effect we expect to see applying treatment $t=1$ to a random individual with observed covariates $X=x$.

    \item In the continuous treatment setting, the \textbf{individual treatment-response curve} (ITRC) for an individual with covariates $x$ is the function 
    \begin{equation}\label{eq:itrc_def}
        t \mapsto \mu_t(x) \qquad t \in \mathcal{T}.
    \end{equation}
    For any value of the treatment, the ITRC maps out the expected outcome for a random individual with covariates $X=x$. The continuous treatment setting lends itself to optimization, where we can ask ``what treatment level maximizes the response?'' The ITRC is also referred to as the individual or conditional \emph{dose-response curve} in medical settings \citep{schwab2020doseresponse}. 
\end{itemize}

\subsection{Data-generating processes and identifiability}\label{subsec:data_generating_processes_and_identifiability}

Let $P = P(X, T, \{Y(t)\}_{t \in \mathcal{T}}, Y)$ denote the joint distribution of the observed covariates, treatment, potential outcomes, and factual outcome. The distribution $P$ is induced by a \emph{data-generating process} (DGP) which specifies how the variables are generated, for example as
\begin{align}\label{eq:example_dgp}
    \begin{aligned}
        X&\sim P_X,\\
        T \mid X &\sim P_{T\mid X},\\
        \{Y(t)\}_{t\in\mathcal{T}} \mid X    &\sim P_{\{Y(t)\}_{t\in\mathcal{T}}\mid X}, \\
        Y &= Y(T).
    \end{aligned}
\end{align}
While the distribution $P$ is fully characterized by the DGP that induces it, $P$ is generally unknown in real world settings. Instead, what we have access to in practice are samples from the \emph{observational distribution} $P_{\text{obs}}$. This is defined as the marginal distribution of $(X, T, Y)$ under $P$, since only the factual outcome, rather than all potential outcomes, can actually be observed:
\begin{equation}\label{eq:obs_dist}
    P_{\text{obs}}(X, T, Y) \coloneqq P(X, T, Y).
\end{equation}
Although $P$ is unknowable in practice, it is a useful abstraction since perfect knowledge of $P$ allows us to compute CEPOs, and thus our other causal estimands of interest. To make this connection, we introduce the do-operator $\operatorname{do}(T=t)$ \citep{pearl2009causality} which represents an intervention that sets $T$ to $t$, as opposed to simply observing $t$ passively. The \emph{interventional distribution} $P(Y \mid \operatorname{do}(T=t))$ is generally not the same as the observational distribution $P_\text{obs}(Y \mid T=t)$, because covariates $X$ or exogenous variables can systematically influence which treatments are observed. For example, if $X$ represents the risk level of a patient, and $t=1$ indicates a more intensive treatment, doctors may only prescribe $t=1$ for high risk patients. In contrast, $\operatorname{do}(T=1)$ forces the intensive treatment, even if the patient is low-risk and doctors would normally not prescribe this treatment. Hence, intervening on treatments without regard to covariates also changes the distribution of outcomes $Y$.

Although the interventional distribution is not the same as $P_\text{obs}(Y \mid T=t)$, it is related to $P$. Under the intervention $\operatorname{do}(T=t)$ the treatment is set to $t$, and by the consistency assumption the resulting outcome is $Y(t)$, which means
\begin{equation}\label{eq:interventional_dist}
    P(Y \mid \operatorname{do}(T=t)) = P(Y(t)).
\end{equation}
\Cref{eq:interventional_dist} shows that the interventional distribution contains rich information on the potential outcomes. By conditioning on $X$ we get the \emph{conditional interventional distribution} (CID) $P(Y \mid \operatorname{do}(T=t), X=x)$, which directly gives the CEPO (\Cref{eq:cepo_def}) via its mean:
\begin{equation}\label{eq:cepo_interventional}
    \int y\, \text{d}P(Y=y \mid \operatorname{do}(T=t), X=x) = \mathbb{E}[Y \mid \operatorname{do}(T=t), X=x] = \mathbb{E}[Y(t) \mid X=x] = \mu_t(x).
\end{equation}
If we instead consider the difference in potential outcomes $Y(1) - Y(0)$, we obtain the \emph{conditional distribution of treatment effects} (CDTE), $P(Y(1) - Y(0) \mid X = x)$. In an analogous manner to \Cref{eq:cepo_interventional}, the CDTE directly gives the CATE via its mean:
\begin{equation}
    \int \Delta \, \text{d}P(Y(1) - Y(0) = \Delta \mid X = x) = E[Y(1) - Y(0) \mid X = x] = \operatorname{CATE}(x),
\end{equation}
where $\Delta$ denotes a realization of the random variable $Y(1) - Y(0)$. 

Note that the joint distribution $P$ or even the CID alone is enough to compute the desirable causal estimands from \Cref{subsec:the-potential-outcomes-framework}. One can sample from the CID if interventions are possible, but often this is not the case. Interventions may be expensive, impractical, or unethical, like giving a patient a treatment which will probably harm them. More often than not we only have observational data, namely samples from $P_\text{obs}$.

Since marginalizing a probability distribution as in \cref{eq:obs_dist} loses information, there are in general many different joint distributions $P$ which could give rise to the same observational distribution $P_{\text{obs}}$. This is linked to the \hyperlink{fundamental_problem_of_causal_inference}{Fundamental Problem of Causal Inference}, now seen as a problem of \emph{identifiability}: understanding the observational distribution is not always enough to identify which DGP it actually derives from. 

Under what conditions \emph{can} we identify a causal effect like the ATE, which is a function of $P$, given access only to $P_{\text{obs}}$? To make this question mathematically precise, let $\mathcal{P}_{\text{DGP}}$ denote the set of joint probability distributions over $(X, T, \{Y(t)\}_{t \in \mathcal{T}}, Y)$, deriving from all DGPs consistent with those variables. Define an equivalence relation on $\mathcal{P}_{\text{DGP}}$ by 
\begin{equation}
    P_1 \sim P_2 \iff (P_1)_{\text{obs}} = (P_2)_{\text{obs}}, \qquad \forall P_1, P_2 \in \mathcal{P}_{\text{DGP}},
\end{equation}
which forms \emph{observational equivalence classes}. Two interventional distributions are \emph{observationally equivalent} if they lie in the same observational equivalence class. Practically, $P_1 \sim P_2$ means that these distributions are indistinguishable given only observational data---no matter how much data is available. 

We define a \emph{causal estimand} to be a functional $g : \mathcal{P}_{\text{DGP}} \to \mathbb{R}$. For example, the ATE (\Cref{eq:ate_def}) is a causal estimand since it is a function of $P$, via potential outcomes, and outputs a scalar.
\begin{definition}\label{def:identifiable}
Let $\mathcal{P} \subset \mathcal{P}_{\text{DGP}}$. A causal estimand $g$ is \textbf{identifiable given} $\mathcal{P}$ if it is constant on observational equivalence classes of $\mathcal{P}$, i.e. for any $P_1, P_2 \in \mathcal{P}$,
\begin{equation}
    (P_1)_{\text{obs}} = (P_2)_{\text{obs}} \implies g(P_1) = g(P_2).
\end{equation}
\end{definition}
In other words, $g$ is identifiable exactly when it can be written as a function of $P_{\text{obs}}$ alone. The subset $\mathcal{P}$ is often clear from context (see below), in which case we say simply that $g$ is identifiable, without explicitly referencing $\mathcal{P}$. Note also that identifiability does not imply that $g$ is computable given a finite dataset sampled from $P_{\text{obs}}$.

Without additional assumptions, the causal estimands we care about from \Cref{subsec:the-potential-outcomes-framework} are not identifiable given $\mathcal{P}_{\text{DGP}}$. The set of possible distributions $\mathcal{P}_{\text{DGP}}$ is simply too large to ensure constancy over all the $P_\text{obs}$ derived from $P\in \mathcal{P}_{\text{DGP}}$. Practitioners must narrow down the allowable set to some $\mathcal{P} \subset \mathcal{P}_{\text{DGP}}$ by making certain assumptions about what DGPs could exist. In this work, we focus on a particular set of identification assumptions commonly used in causal inference. More generally, Pearl's do-calculus \citep{pear1995causaldiagrams} gives a complete set of graph-theoretic rules for identification: if a causal estimand is identifiable, do-calculus provides a method for calculating it in terms of $P_{\text{obs}}$ \citep{shpitser2006identification, huang2012pearlscalculusinterventioncomplete}. 

The most common set of assumptions used in the field of causal inference in order to guarantee identifiability are the following \citep{imbens2015causalinference, neal2020introduction, yao2021surveyoncausalinference, balazadeh2026iviclboundingcausaleffects}.

\begin{definition}
A DGP satisfies the \textbf{ignorability} assumption if for all $t \in \mathcal{T}$,
\begin{equation}\label{eq:ignorability_def}
    Y(t) \independent T \mid X.
\end{equation}
This is also called the (conditional) unconfoundedness assumption. It says that conditional on $X$, treatment assignment is independent of the potential outcomes. Individuals who actually received treatment $t$ are representative, with respect to $Y(t)$, of individuals who could have received treatment $t$. This allows us to identify the distribution of potential outcomes from observed factual outcomes because ignorability implies
\begin{equation}\label{eq:ignorability_example}
    P(Y(t) \mid X = x) = P(Y \mid T=t, X=x).
\end{equation}
\end{definition}
\begin{definition}\label{def:postivity}
In the binary or multi-armed treatment case, a DGP satisfies the \textbf{positivity} assumption if 
\begin{equation}\label{eq:positivity_binary_or_multiarm_def}
    P(T = t \mid X = x) > 0 \qquad \forall t \in \mathcal{T}, x \in \mathcal{X}.
\end{equation}
In the continuous treatment case, a DGP satisfies the positivity assumption if the conditional density function $p$ for $P$ satisfies
\begin{equation}\label{eq:positivity_continuous_def}
    p(T = t \mid X = x) > 0 \qquad \text{almost everywhere}.
\end{equation}
The positivity assumption is also called the overlap assumption. It is needed because we can only learn the causal effect of a treatment for covariate values where that treatment can actually be observed. For example, if we assume ignorability, then by \cref{eq:ignorability_example} the quantity $\mathbb{E}[Y(t) \mid X=x]$ is equal to $\mathbb{E}[Y \mid T=t, X=x]$, but if $P(T = t \mid X = x) = 0$ there will be no observations of $Y$ for individuals with covariates $X=x$ who receive treatment $T=t$.
\end{definition}
If a DGP satisfies both positivity and ignorability, it is said to satisfy \emph{strong ignorability}.
\begin{definition}\label{def:sutva}
A DGP satisfies the \textbf{stable unit treatment value assumption} (SUTVA) if: 
\begin{enumerate}
    \item (No interference) The individual-level potential outcome $Y_n(t_n)$ for individual $n$ does not depend on the treatment assigned to any other individual $m$, and;
    \item (No hidden versions of treatments) The treatment $t$ is a well-defined intervention without multiple versions that have different effects.
\end{enumerate}

Without the first part of the SUTVA, $Y_n(t_n)$ is not well-defined from $t_n$ alone; $Y_n(t_n)$ would instead need to express dependence on other treatment assignments as $Y_n(t_1, ..., t_n, ..., t_{N})$. No interference implies $Y_n(t_1, ..., t_n, ..., t_{N}) = Y_n(t_n)$. Without the second part there are different versions of $t_n$ that produce different outcomes, so $Y_n(t_n)$ does not uniquely specify an outcome. The SUTVA ensures that the potential outcome $Y_n(t_n)$ is a well-defined object which can then be linked to the corresponding factual outcome $Y_n$ under $t_n$ via consistency.
\end{definition}

Together, these three assumptions guarantee that the CID (and hence the CEPO) is identifiable, and can be recovered from $P_{\text{obs}}$ via the \emph{backdoor adjustment} \citep{pearl2009causality, neal2020introduction}
\begin{equation}
    P(Y \mid \operatorname{do}(T=t)) = \int P(Y \mid T=t, X=x) P(X=x) \, \text{d}x.
\end{equation}
Accordingly, the three assumptions together are commonly referred to as the backdoor setting. In terms of the mathematical framework discussed above, let $\mathcal{P}_{\text{back}} \subset \mathcal{P}_{\text{DGP}}$ be defined by 
\begin{equation}\label{eq:p_back}
    \mathcal{P}_{\text{back}} = \big\{ P \in \mathcal{P}_{\text{DGP}} \mid P \ \text{satisfies strong ignorability and SUTVA} \big\}.
\end{equation}
Then the CID and CEPO are identifiable given $\mathcal{P}_{\text{back}}$. The main drawback of the backdoor setting is that ignorability is an untestable assumption in practice \citep{pearl2009causality}.

A well-known setting in which we have partial identifiability is the \emph{instrumental variable} (IV) setting \citep{manski2003partial}.\footnote{The IV setting also implies \emph{parametric} identifiability; that is, identifiability when restricting to a narrow subclass of interventional distributions, e.g. those arising from linear models \citep{neal2020introduction}.} While we do not give a formal definition here, partial identifiability occurs when the causal estimand of interest cannot be determined exactly, but its range is reduced to a finite interval. An instrumental variable $Z$ is one which has a causal effect on the outcome $Y$ which is fully mediated by the treatment $T$, and for which there are no unobserved confounders between $Z$ and $T$ or $Y$ \citep[Chapter~9]{neal2020introduction}. 

Finally, we mention that the \emph{frontdoor setting} is another set of assumptions for which identification holds \citep[Chapter~6]{neal2020introduction}. However, this setting is much less used in practice \citep{imbens2020potentialoutcomeanddirectedacyclicgraph}.

\subsection{Bayesian networks and structural causal models}\label{subsec:scms}

We have discussed data-generating processes and identifiability of causal effects. In this section, we discuss a well-known graph-theoretic framework for quantitatively describing DGPs. This framework also plays a key role in specifying priors for CFM training (\Cref{subsec:training_cfms}).

Let $G$ be a directed acyclic graph (DAG). For a node $v \in G$, we let $\text{pa}(v)$ denote the set of all parents of $v$, that is, the set of all nodes $u \in G$ for which there is an edge $u \to v$. When the nodes of a DAG represent random variables and the edges represent dependence relationships between these random variables, we call the DAG a \emph{Bayesian network} \citep{pearl2009causality}. In this work, we mainly consider the case where edges represent direct causal relationships, in which case the Bayesian network is sometimes referred to as a \emph{causal network}.

Bayesian networks are useful for describing conditional dependencies between random variables. If $Z_1, \ldots, Z_K$ are random variables, we are often interested in the joint probability distribution 
\begin{align*}
    P(Z_1, \ldots, Z_K) &= P(Z_1)\prod_{k = 2}^K P(Z_k \mid Z_1, \ldots, Z_{k - 1}). 
\end{align*}
Bayesian networks let us factor the right-hand side into a much sparser product by conditioning each variable only on its parents:
\begin{equation}
    P(Z_1, \ldots, Z_K) = \prod_{k} P\big(Z_k \mid \text{pa}(Z_k)\big).
\end{equation}
Bayesian networks describe dependencies or sometimes causal relationships between variables, but not how to compute causal effects. Augmenting a Bayesian network with explicit equations relating each node gives a model that can be used to simulate data and perform interventions on.

\begin{definition}
A \textbf{structural causal model} (SCM) is a Bayesian network $G$ of $K$ nodes, together with a set of \emph{exogenous variables} $\{U_k\}_{k = 1}^K$ and a set of functions $\{f_v\}_{v \in G}$ such that
\begin{equation}\label{eq:def_structural_equations}
    Z_k = f_{Z_k}(\text{pa}(Z_k), U_k) \qquad \forall \ k = 1, \ldots, K.
\end{equation}
\end{definition}
The variables $Z_k$ are called \emph{endogenous variables} and represent the modeled set of variables (including covariates $X$, treatments $T$, and potential outcomes $Y$), whereas the $U_k$ represent unmodeled factors or noise. The equations \ref{eq:def_structural_equations} are called the \emph{structural equations} of the SCM; see \Cref{ex:scm-ex-1}. Note that the structural equations only model the endogenous variables $Z_k$, not the exogenous $U_k$. We will denote SCMs using the letter $S$.

Given a realization of the exogenous variables $U_k$, we can simulate the entire system by passing these values into the structural equations, starting with nodes in the graph that do not have parents and progressing according to the DAG structure. We typically posit a distribution $P_U$ over the exogenous variables, allowing us to sample $U_k \sim P_U$, and then apply the structural equations to simulate observational data. SCMs also enable us to model factual, counterfactual, and interventional scenarios. For example, the intervention $\operatorname{do}(T = t)$ corresponds to replacing the structural equation 
\begin{align*}
    T &= f_T(\text{pa}(T), U_T)
\end{align*}
with $T = t$. We can then sample $U_k \sim P_U$ to approximate the CID $P(Y \mid \operatorname{do}(T = t), X = x)$, or fix the values of each $U_k$ while intervening on $T$ to generate counterfactuals $Y(t)$. SCMs are therefore examples of DGPs; we will see the data-generative capabilities of SCMs used to great effect in \Cref{sec:cfms-core}.

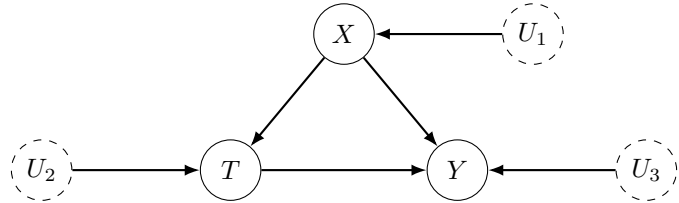
\begin{figure}[ht]
    \centering
\begin{tikzpicture}[
    every node/.style={circle, draw, minimum size=8mm, inner sep=0pt}
]
    \node (x) at (0, 1.8) {$X$};
    \node (t) at (-1.5, 0) {$T$};
    \node (y) at (1.5, 0) {$Y$};

    \node[dashed] (u1) at (2.5, 1.8) {$U_1$};
    \node[dashed] (u2) at (-4, 0) {$U_2$};
    \node[dashed] (u3) at (4, 0) {$U_3$};

    \draw[-{Latex[length=2mm]}, thick] (x) -- (t);
    \draw[-{Latex[length=2mm]}, thick] (x) -- (y);
    \draw[-{Latex[length=2mm]}, thick] (t) -- (y);
    \draw[-{Latex[length=2mm]}, thick] (u1) -- (x);
    \draw[-{Latex[length=2mm]}, thick] (u2) -- (t);
    \draw[-{Latex[length=2mm]}, thick] (u3) -- (y);
\end{tikzpicture}
    \caption{An example DAG arising from the SCM given in Example \ref{ex:scm-ex-1}, with endogenous variables $\{X, T, Y\}$ and exogenous variables $\{U_1, U_2, U_3\}$. Dashed nodes denote exogenous variables.}
    \label{fig:scm-ex-1}
\end{figure}

\begin{example}\label{ex:scm-ex-1}
Consider endogenous variables $\{X, T, Y\}$, exogenous variables $\{U_1, U_2, U_3\}$, and structural equations
\begin{align*}
    x &= f_X(u_1) \coloneqq u_1, \\ 
    t &= f_T(x, u_2) \coloneqq x^2 + 1 + u_2, \\ 
    y &= f_Y(x, t, u_3) \coloneqq t - x + u_3,
\end{align*}
with $U_1, U_2, U_3 \overset{\text{i.i.d.}}{\sim} \mathcal{N}(0, 1)$. The corresponding DAG is shown in \Cref{fig:scm-ex-1}. Sampling $u_1, u_2, u_3$ and propagating through $f_X, f_T, f_Y$ generates i.i.d. draws from the observational distribution $P_\text{obs}(X, T, Y)$. To generate interventional data, we simply replace $f_T$ with a fixed constant $t_0$ during the propagation, simulating the effect of the intervention $\operatorname{do}(T = t_0)$.
\end{example}

\subsection{Bayesian inference and posterior predictive distributions}\label{subsec:bayesian_inference_ppd}

Let us revisit the set $\mathcal{P}_\text{DGP}$ of joint probability distributions from \Cref{subsec:data_generating_processes_and_identifiability}, and denote a specific DGP by a parameter $\psi$, writing $P^\psi\in \mathcal{P}_\text{DGP}$ for its joint distribution with observational marginal $P^\psi_{\text{obs}}$. 
Observational datasets $\mathcal{D}_{\text{obs}} = \{(x_n, t_n, y_n)\}_{n = 1}^N$ are drawn i.i.d. from $P^\psi_{\text{obs}}$.

Given only observational data $\mathcal{D}_{\text{obs}}$, we would like to estimate a causal estimand $g$. As we learned in \Cref{subsec:data_generating_processes_and_identifiability}, causal estimands can be computed from $P^\psi$, but require identifiability conditions to be computable from $P^\psi_{\text{obs}}$. Even under identifiability conditions, the finite sample available in $\mathcal{D}_{\text{obs}}$ leaves uncertainty about which DGP the data came from. Hence, estimating $g$ needs to incorporate information about several possible DGPs. The standard way to proceed via Bayesian inference is to assume a prior $\pi(\psi)$ over candidate DGPs and use the observational dataset $\mathcal{D}_{\text{obs}}$ to refine that distribution into a posterior $\pi(\psi \mid \mathcal{D}_{\text{obs}})$ via Bayes' rule. We can then express our knowledge of the causal estimand $g$ through its \emph{posterior predictive distribution} (PPD), defined as 
\begin{equation}\label{eq:ppd_def}
    \pi^g( [a, b] \mid \mathcal{D}_{\text{obs}} ) = \int \pi(\psi \mid \mathcal{D}_{\text{obs}})\bm{1}_{g(P^\psi) \in [a, b]} \, \text{d}\psi,
\end{equation}
for any $a, b \in \mathbb{R}$ with $a \leq b$. The PPD expresses our uncertainty about the value of $g$, including from observing a finite sample of data, and from our uncertainty about which DGP generated the data. For any interval $I \subset \mathbb{R}$, for example $[-1, 1]$, this distribution gives us the posterior probability that the value of $g$ is in $I$. For any $\psi$, $g(P^\psi)$ takes on a single value; for any $I$, \cref{eq:ppd_def} adds up the posterior probability of all DGPs $\psi$ for which the value $g(P^\psi)$ is consistent with the specified range $I$.

\begin{example} When $g(P^\psi)$ is the CEPO $\mu_t(x; P^\psi) = \mathbb{E}_{P^\psi}[Y(t) \mid X = x]$, the CEPO-PPD is 
\begin{equation*}
    \pi^{\mu_t}([a, b] \mid x, \mathcal{D}_{\text{obs}}) = \int \pi(\psi \mid \mathcal{D}_{\text{obs}})\bm{1}_{\mu_t(x; P^\psi) \in [a, b]}\, \text{d}\psi.
\end{equation*}
\end{example}

\begin{example} When $g(P^\psi)$ is the ATE, given by $\mathbb{E}_{P^\psi}[Y(1) - Y(0)]$, the ATE-PPD is 
\begin{equation*}
    \pi^{\text{ATE}}([a, b] \mid \mathcal{D}_{\text{obs}}) = \int  \pi(\psi \mid \mathcal{D}_{\text{obs}}) \bm{1}_{\text{ATE}(P^\psi) \in [a, b]} \, \text{d}\psi.
\end{equation*}
\end{example}

Sometimes directly obtaining causal estimands is not the goal, and more benefit is gained by modeling the CID itself. The CID can also be formulated as a PPD to enable its estimation from observational data. Once again we assume a prior over DGPs which becomes a posterior after observing data $\mathcal{D}_{\text{obs}}$. When we further condition the CID on observed data, we can express our uncertainty about which DGP generated the data via the CID-PPD:
\begin{equation}\label{eq:cid-ppd}
    P(Y \mid \operatorname{do}(T=t), X=x, \mathcal{D}_{\text{obs}}) = \int P(Y \mid \operatorname{do}(T=t), X=x, \psi)\pi(\psi \mid \mathcal{D}_{\text{obs}}) \, \text{d}\psi.
\end{equation}
The CID-PPD can be understood as separating two types of uncertainty. First, for the DGP posterior $\pi(\psi \mid \mathcal{D}_{\text{obs}})$, observing enough data under identifiability conditions would in principle collapse $\pi(\psi \mid \mathcal{D}_{\text{obs}})$ onto one DGP. Hence, the DGP posterior represents \emph{epistemic} uncertainty as to which DGP generated the data we observed due to our lack of knowledge. If identifiability assumptions do not hold, the DGP posterior would also represent epistemic uncertainty from our inability to distinguish observationally equivalent DGPs. This uncertainty is structural, and would only be reducible by performing interventions. On the other hand, $P(Y \mid \operatorname{do}(T=t), X=x, \psi)$ represents \emph{aleatoric} uncertainty about which outcome we will observe under an intervention, assuming a fixed DGP $\psi$. This uncertainty arises from randomness in the DGP, including any noise and exogenous variables that are not observed.

If one is mainly interested in treatment effect estimates such as the CATE or ATE, a natural object to model is the CDTE. Similarly to the CID, this can also be formulated as a PPD by conditioning on observed data. The CDTE-PPD is:
\begin{equation}\label{eq:cdte-ppd}
    P(Y(1) - Y(0) \mid X = x, \mathcal{D}_{\text{obs}}) = \int P(Y(1) - Y(0) \mid X = x, \psi) \pi(\psi \mid \mathcal{D}_{\text{obs}})\, \text{d}\psi.
\end{equation}
This is similar to the PPD but targets only the difference in potential outcomes rather than each potential outcome individually.

The PPD is a fundamental object in Bayesian inference and data analysis \citep{gelman2013bayesiandataanalysis} and is not specific to causal inference, yet it has no closed-form solution in general and must be approximated. Popular traditional methods involve first approximating the posterior $\pi (\psi \mid \mathcal{D}_{\text{obs}})$ and then using eqs. (\ref{eq:ppd_def})--(\ref{eq:cdte-ppd}) to calculate the PPD. Many techniques have historically been developed for posterior approximation, such as Markov Chain Monte Carlo (MCMC) methods \citep{neal1996bayesian, hoffman2014nouturn}, variational inference \citep{jordan1999introduction, wainwright2008graphical}, and neural posterior estimation \citep{papamakarios2016fastepsilonfree}.

\subsection{Prior-data fitted networks}\label{subsec:pfns}

More recently, \emph{prior-data fitted networks} (PFNs) \citep{muller2022transformers} have been introduced as transformers \citep{vaswani2017attention} that can perform Bayesian inference by leveraging in-context learning \citep{brown2020icl}. Moreover, contrary to the traditional methods of Bayesian inference, PFNs go directly from dataset to PPD without explicitly approximating the posterior. A valuable consequence of this is that PPDs provide uncertainty quantification in the same forward pass used for prediction \citep{muller2022transformers, balazadeh2026causalpfn, melnychuk2026frequentistconsistencypriordatafitted}.

PFNs were first developed as a model of a generic PPD. Suppose there is an underlying set of possible tasks $\varphi$ over which we assume some prior $p(\varphi)$. Given a specific supervised predictive (not causal) task $\varphi$, a finite labeled dataset $\mathcal{D}_\text{sup} = \{ (x_n, y_n)\}_{n=1}^N$ is sampled. Then after defining the posterior $p(\varphi\mid \mathcal{D}_\text{sup})$, the PPD can be expressed as
\begin{equation}\label{eq:generic_ppd}
    p(y\mid x, \mathcal{D}_\text{sup}) = \int p(y\mid x, \varphi)p(\varphi\mid \mathcal{D}_\text{sup})  \, \text{d}\varphi.
\end{equation}
This PPD reflects our lack of knowledge about which task was used to sample $\mathcal{D}_\text{sup}$. Given the supervised dataset $\mathcal{D}_\text{sup}$, the PPD can be used to estimate the label distribution of a new datapoint $x$ drawn from the same task $\varphi$.

Instead of approximating the posterior, PFNs learn a mapping from a dataset of samples to the PPD directly, which is generically called \emph{amortized Bayesian inference}. At inference, they take an entire supervised dataset as context, and for each query point $x$ they directly output $q_\theta(y\mid x, \mathcal{D}_\text{sup})$, a parametrized model of $p(y\mid x, \mathcal{D}_\text{sup})$. This powerful paradigm is enabled by three factors:
\begin{enumerate}
    \item The \emph{prior-data loss} which minimizes the cross entropy between $p(y\mid x, \mathcal{D}_\text{sup})$ and $q_\theta(y\mid x, \mathcal{D}_\text{sup})$, without needing to evaluate $p(y\mid x, \mathcal{D}_\text{sup})$;
    \item Tractable yet highly diverse priors $p(\varphi)$ that generate unlimited synthetic tasks and datasets;
    \item A transformer-based architecture enabling in-context learning.

\end{enumerate}

First, the prior-data loss is defined as the negative log-likelihood of the PFN's output distribution $q_\theta$ at the correct label $y$, when given the corresponding $x$ and supervised dataset as context:
\begin{equation}\label{eq:predictive_data_prior_loss}
    \mathcal{L}_{\text{pred}}(\theta) = \mathbb{E}_{\varphi \sim p(\varphi), \, \mathcal{D}_{\text{sup}} \cup \{x, y\} \sim P^\varphi}\big[ -\log( q_\theta(y \mid x, \mathcal{D}_{\text{sup}}) ) \big].
\end{equation}
This loss is tractable because it only requires evaluating the model $q_\theta$ at the value of the true label, and does not require evaluating the true PPD. Nevertheless, a brief computation \citep{muller2022transformers} shows that this loss is equal to the forward KL divergence between the true PPD and $q_\theta$. Thus, minimizing the prior-data loss encourages $q_\theta$ to be a good approximation to the PPD. In practice, model parameters $\theta$ are updated using stochastic gradient descent (SGD).

 Second, notice that SGD training with the prior-data loss $\mathcal{L}_{\text{pred}}(\theta)$ requires sampling tasks $\varphi\sim p(\varphi)$, then sampling supervised datasets $\mathcal{D}_\text{sup}$. This could be accomplished by collecting relevant real-world datasets corresponding to various tasks, and then iterating over them for gradient steps. However, when pre-training a PFN it is not assumed that one knows what task $\varphi$ will need to be solved at inference. To be true foundation models, PFNs are expected to generalize to \emph{any} unseen task at inference. This requires that the prior $p(\varphi)$ have as broad coverage as possible. In practice most PFNs have relied on synthetic priors which are tractable to sample from and produce highly diverse datasets. Training proceeds by sampling new prior-synthetic data each batch, exposing the PFN to maximally diverse data over the course of training.

Third, a model architecture to support direct PPD prediction via $q_\theta(y \mid x, \mathcal{D}_{\text{sup}})$ must accept an entire dataset at inference to facilitate predictions for $x$. Modern foundation models often do not require training on the specific inference task because they leverage in-context learning. This is in turn enabled by the attention mechanism of transformers, which allows query tokens (embeddings of $x$) to attend to context tokens (embeddings of the labeled examples in $\mathcal{D}_\text{sup}$).

In summary, a PFN is pretrained using the prior-data loss with synthetic data sampled from a prior. At inference, it leverages in-context learning to predict the label distribution (PPD) of query data using supervised samples from a never-before-seen task, doing so without updates to the PFN's weights.

Early demonstrations of PFNs showed strong performance in a variety of areas \citep{muller2022transformers}, and quickly excelled at tabular predictive tasks  \citep{hollmann2023tabpfn, hollmann2025accurate, ma2025tabdpt, qu2025tabicl}, marking the advent of tabular foundation models (TFMs). In this setting each $x$ is a set of numerical features, and $y$ is a classification or regression label. PFNs have had similar success in relational learning \citep{wang2026relationalincontextlearningsynthetic, hayler2026advancingopenreproduciblerelational} and time-series forecasting \citep{dooley2023forecastpfn, taga2025timepfn}, where the data structures are somewhat more complex, but it is still possible to design synthetic data priors.

\section{Causal Foundation Models}\label{sec:cfms-core}

In \Cref{subsec:bayesian_inference_ppd} we arrived at the theoretical underpinning of modeling causal estimands and distributions from observational data: assuming a prior over DGPs and identifiability conditions, the PPD of a causal estimand $g$ (\Cref{eq:ppd_def}) gives a distribution over the likely values that $g$ takes, accounting for uncertainty from our finite observations, and from our lack of knowledge of which DGP generated the data. Similar observations held for the CID-PPD (\Cref{eq:cid-ppd}) and CDTE-PPD (\Cref{eq:cdte-ppd}). Then, in \Cref{subsec:pfns} we saw that PFNs have emerged as a powerful paradigm for estimating generic PPDs, and have been widely used for predictive tabular tasks. It is natural to ask how PFNs might be applied to the relevant PPDs from causal inference. This connection is the basis for causal foundation models.

\subsection{What are CFMs?}

Causal foundation models use PFNs to amortize Bayesian inference for causal quantities like the CEPO or the CID. Though the field is quickly evolving, core qualities of CFMs have gradually emerged and solidified. We therefore adopt the following as our working definition of a causal foundation model. 

\begin{tcolorbox}[colback=teal!5, colframe=teal!60!black, title=Causal Foundation Model]
\begin{definition}\label{def:cfm} A \textbf{causal foundation model (CFM)} is a prior-data fitted network that has been pretrained on a variety of causal tasks in order to estimate causal quantities from new datasets using in-context learning. We emphasize that a CFM predicts on unseen tasks without updating any model weights. 
\end{definition}
\end{tcolorbox}

CFMs apply the large-scale prior-fitting methodology of PFNs to causal inference. During pretraining, the CFM is repeatedly shown a labeled observational dataset generated by an SCM sampled from a prior, together with a causal query which can be computed from the SCM, but which must be estimated by the model from the observational dataset alone. Through many such iterations, the CFM learns a mapping from an observational dataset to a PPD for the causal quantity. At deployment, the weights remain fixed. Causal effect estimation occurs through in-context learning \citep{brown2020icl} based on examples from an observational dataset, rather than weight updates.

\begin{figure}[t]
    \centering
    \includegraphics[width=0.8\linewidth]{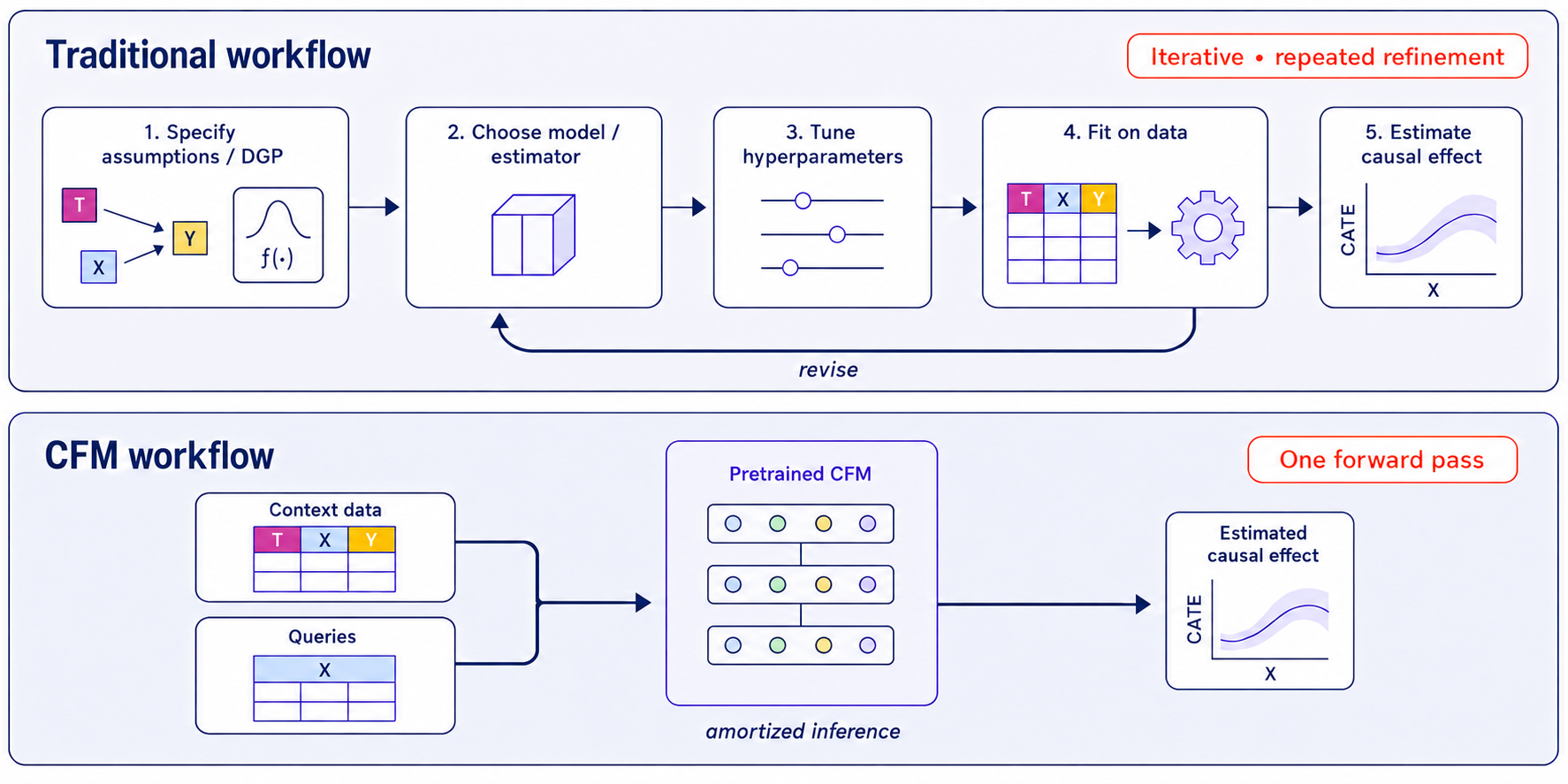}
    \caption{Traditional vs. CFM workflows. \textit{(Top)} For each new problem, the traditional pipeline involves analyzing the dataset, proposing a model, and training it before inference. \textit{(Bottom)} CFMs are applied off-the-shelf to new problems, without training, fine-tuning, or hyperparameter optimization. The reader is invited to experiment with several implemented CFMs for themselves \jupyter{https://github.com/layer6ai-labs/cfms/blob/main/notebooks/Foundation_models_sandbox.ipynb}.
}
    \label{fig:workflow-difference}
\end{figure}

CFMs vastly speed up the workflow for causal effect estimation. A traditional workflow requires the analyst to understand a dataset and the causal connections between covariates, specify a causal estimand, choose an adjustment set or identification strategy, fit nuisance models, diagnose overlap, and compute an estimator. A CFM compresses much of this workflow into a single forward pass of a pretrained network. This can greatly reduce friction for non-expert users and enable rapid causal analysis across many small datasets (\Cref{fig:workflow-difference}).

Moreover, they have demonstrated top results on causal tasks, performing above or on par with leading causal inference models which are trained for each task (see our benchmarking in \Cref{sec:benchmarking}). They also perform better than TFMs applied off-the-shelf to causal problems, demonstrating the effectiveness of pretraining on an intrinsically causal prior \citep{robertson2025do-pfn, balazadeh2026causalpfn}. We discuss the pretraining process of CFMs further in \Cref{subsec:training_cfms}.

The first CFMs that fit the above definition were proposed simultaneously by \citet{robertson2025do-pfn} and \citet{balazadeh2026causalpfn}. Focusing on the binary treatments setting, Do-PFN \citep{robertson2025do-pfn} models the CID-PPD, while CausalPFN \citep{balazadeh2026causalpfn} models the CEPO-PPD, among other differences in design choices. Shortly after, CausalFM \citep{ma2026causalfm} was introduced which models the CDTE-PPD but aims to cover the IV and frontdoor settings in addition to backdoor (see \Cref{subsec:data_generating_processes_and_identifiability}). We provide implementations of these CFMs for the reader to experiment with \jupyter{https://github.com/layer6ai-labs/cfms/blob/main/notebooks/Foundation_models_sandbox.ipynb}. Another related line of work estimates marginal interventional distributions using transformer-based neural processes \citep{dhir2025estimatinginterventionaldistributions}.

Although we have framed PFNs as a perfect fit for modeling causal quantities like the CEPO-PPD or CID-PPD, there are several adaptations that are needed to bring PFNs from the world of predictive tasks into the world of causal inference. Recall from \Cref{subsec:pfns} the three enabling factors introduced by PFNs: the prior-data loss $\mathcal{L}_{\text{pred}}(\theta)$, synthetic and tractable priors $p(\varphi)$, and transformer-based architectures for in-context learning (ICL). In the following, we will discuss how each of these factors must be updated to construct a CFM.

\subsubsection{The causal prior-data loss}

The prior-data loss in \cref{eq:predictive_data_prior_loss} supports training a PFN to map directly from a supervised dataset to a PPD over predictive labels. CFMs are trained using a modified version of the prior-data loss called the \emph{causal prior-data loss} \citep{balazadeh2026causalpfn}. The exact form of the causal prior-data loss depends on which PPD is being targeted, but all share an important distinction from the prior-data loss for predictive PFNs. In \cref{eq:predictive_data_prior_loss} the model must predict a distribution over labels $y$ given features $x$, and is shown examples in the same format, namely $(x, y)$ pairs via $\mathcal{D}_{\text{sup}}$. In contrast, we want CFMs to solve a \emph{causal} task from purely observational data; even though the CFM must predict a causal quantity for any covariate-treatment pair, the examples it is shown are of the form $(x, t, y)$ via $\mathcal{D}_{\text{obs}}$ (a single treatment and factual outcome per covariate set), \emph{not} of the form $(x, t_0, y(0), t_1, y(1))$, nor any similar set like $(x, t_0, \mu_{t_0}(x), t_1, \mu_{t_1}(x))$. These latter options contain interventional information which would be useful for learning causal PPDs, but are not feasible to obtain at inference time, so they also should not be used for training. Due to this mismatch of observations and predictions inherent to causal inference, CFMs are trained on a structurally harder task than predictive PFNs.

For the CEPO-PPD (recall the notation from \Cref{subsec:bayesian_inference_ppd}), the causal prior-data loss takes the form
\begin{equation}\label{eq:cepo_data_prior_loss}
    \mathcal{L}_t(\theta) = \mathbb{E}_{\psi \sim \pi, \, \mathcal{D}_{\text{obs}} \cup \{x\} \sim P^\psi_{\text{obs}}}\big[ -\log(q_\theta( \mu_t(x; P^\psi) \mid x, t, \mathcal{D}_{\text{obs}})) \big].
\end{equation}
To compute this loss, a DGP is first sampled from the prior over DGPs $\pi(\psi)$ specified by the developer; then, an observational dataset $\mathcal{D}_{\text{obs}}$ and an additional set of query covariates $x$ are sampled according to $P^\psi_{\text{obs}}$. The ground-truth value $\mu_t(x; P^\psi)$ is computed from the sampled DGP $\psi$, and we evaluate the model's likelihood assigned to that value.

Similarly, for the CID-PPD the loss takes the form
\begin{equation}\label{eq:counterfactual_data_prior_loss}
    \mathcal{L}_t(\theta) = \mathbb{E}_{\psi \sim \pi,\, \mathcal{D}_{\text{obs}} \cup \{x\} \sim P^\psi_{\text{obs}},\, y \sim P^\psi(\cdot\mid \text{do}(T = t), x)} \big[ -\log( q_\theta ( y \mid x, t, \mathcal{D}_{\text{obs}} ) ) \big].
\end{equation}
Again, a DGP is sampled and used to generate an observational dataset and query datapoint from $P^\psi_{\text{obs}}$. The treatment $t$ is applied as an intervention on $x$ to get the outcome $y$, sampled from the true CID. Then the loss evaluates the model's likelihood assigned to this outcome.

For both losses, note that in practice a sum over $\mathcal{L}_t(\theta)$ for all $t\in\mathcal{T}$ is used; for a single query $x$ the model must accurately produce the CEPO-PPD/CID-PPD for any $t$.
Furthermore, both of these losses are tractable as long as interventional data can be simulated. They require evaluating the model $q_\theta$ and accessing ground-truth causal quantities, namely the CEPO value $\mu_t(x; P^\psi)$ or outcome $y(t; P^\psi)$. Importantly, they do not require knowing the true underlying PPDs in closed form or evaluating their likelihoods, which would greatly limit the set of DGPs we could practically use in the prior $\pi(\psi)$. In \Cref{subsec:training_cfms} we return to the point of how ground-truth causal quantities are simulated in practice.

Training with the causal prior-data losses also converges to the true PPD under suitable conditions. \citet{balazadeh2026causalpfn} show that, assuming almost all $\psi$ in the support of $\pi(\psi)$ satisfy positivity, the CEPO prior-data loss in \cref{eq:cepo_data_prior_loss} is equal to the expected forward KL divergence between the true CEPO-PPD and $q_\theta$, up to a constant. This means that training under the causal prior-data loss will in principle converge to the true CEPO-PPD. To recover a point estimate of the CEPO, CausalPFN returns the mean $\mathbb{E}[q_\theta(\mu_{t}(x) \mid x, t, \mathcal{D}_{\text{obs}})]$. Crucially, \citet{balazadeh2026causalpfn} prove that as the size of $|\mathcal{D}_{\text{obs}}|$ grows, this mean converges to the true CEPO if and only if the CEPO is identifiable given the support of the prior $\pi$.

Similarly, for the CID prior-data loss, \citet{robertson2025do-pfn} show that training with \cref{eq:counterfactual_data_prior_loss} minimizes the expected forward KL divergence between the true CID-PPD and $q_\theta$. Thus, training with the causal prior-data loss yields a good approximation $q_\theta$ of the causal target. However, this does not make all causal effects estimatable. For example, if the desired causal estimands are not identifiable given the support of the prior $\pi$, then it is impossible to predict an accurate point estimate, even with infinite data. Non-identifiability implies that the posterior distribution $P(\psi \mid \mathcal{D}_{\text{obs}})$ does not collapse around a single value $\psi$, as multiple DGPs could explain the observed data. As a consequence, non-identifiability would be expressed as uncertainty in the CID.

\subsubsection{Tractable synthetic priors over DGPs}\label{sec:synthetic_priors}

The next step in applying the machinery of \Cref{subsec:pfns} to causal inference is to design causal priors from which both observational and interventional data can be sampled. SCM-based priors make this possible (\Cref{subsec:scms}).

Synthetic data priors have been adopted for predictive PFNs, including TFMs, because collecting real-world tabular data at scale and with sufficient diversity has proved challenging. However, synthetic data is not necessary for TFMs---real data can be used, as done for example by TabDPT \citep{ma2025tabdpt}, ConTextTab \citep{spinaci2025contexttab}, and RealTabPFN \citep{garg2025real}, because the model $q_\theta$ only needs to be evaluated at the label $y$ to compute the prior-data loss (\Cref{eq:generic_ppd}). In contrast we saw that eqs.~\ref{eq:cepo_data_prior_loss} and \ref{eq:counterfactual_data_prior_loss} require evaluating the model at the true CEPO value, or an interventional outcome, respectively. These causal quantities are not available in real observational data, which makes synthetic priors more of a necessity, not merely a convenience. Furthermore, we can never obtain counterfactual information for any individual. For coarsely-defined covariate strata, an experiment may contain comparable individuals exposed to different treatments; however, as the resolution of the strata increases (and thus becomes more specific), finding comparable individuals becomes more rare. A causal prior-data loss requires reliable ground-truth labels for any combination of covariates and intervention, which is simply not available in real data at sufficient scale and diversity to train a CFM.

Synthetic priors have therefore been a fundamental and scalable ingredient to all current CFMs, able to generate a highly diverse and high-quality sample of tasks for pretraining. However, an added complexity is the question of identifiability. When designing the prior, one must decide which subset $\mathcal{P} \subset \mathcal{P}_{\text{DGP}}$ of DGPs the prior covers. 

\begin{table}[t]
\centering
\small
\caption{The main design choices of the first CFMs. All are PFNs and use in-context learning, but differ in their prediction target, prior design, identification strategy, and architectural details.}
\label{tab:first-cfms}
\setlength{\tabcolsep}{3pt}
\begin{tabular}{@{}p{0.14\linewidth}p{0.27\linewidth}p{0.24\linewidth}p{0.27\linewidth}@{}}
\toprule
{\raggedright Model\par} & {\raggedright Prediction target\par} & {\raggedright Prior and identification\par} & {\raggedright Embedding and representation\par} \\
\midrule
{\raggedright Do-PFN \citep{robertson2025do-pfn}\par}
& {\raggedright CID-PPD \mbox{$P(Y \mid \operatorname{do}(T = t), X = x, \mathcal{D}_{\text{obs}})$}\par}
& {\raggedright Non-identifiable prior.\par}
& {\raggedright {TabPFNv1-style transformer with a treatment-column indicator.}\par} \\
\addlinespace
{\raggedright CausalPFN \citep{balazadeh2026causalpfn}\par}
& {\raggedright CEPO-PPD \mbox{$\pi^{\mu_t}(\mu_t \mid X = x, \mathcal{D}_{\text{obs}})$}\par}
& {\raggedright Identifiable prior via strong ignorability (backdoor setting).\par}
& {\raggedright {Treatment is concatenated with covariates before embedding; outcome is embedded separately. }\par} \\
\addlinespace
{\raggedright CausalFM \citep{ma2026causalfm}\par}
& {\raggedright CDTE-PPD \mbox{$P(Y(1) - Y(0) \mid X = x, \mathcal{D}_{\text{obs}})$}\par}
& {\raggedright Separate priors and models for backdoor, frontdoor, and instrumental variable settings. \par}
& {\raggedright {Separately encodes covariates, treatment, and outcome; transformer output passed through GMM head.}\par} \\
\bottomrule
\end{tabular}
\end{table}

Do-PFN employs a non-identifiable prior, meaning that their prior may generate DGPs having different CID-PPDs but the same observational distributions. The motivation is to enable their model to report an increase in uncertainty coming from non-identifiability, for example, in the presence of unobserved confounders \citep{robertson2025do-pfn}. On the other hand, CausalPFN is built with a backdoor prior, meaning that its prior is supported on the set of DGPs $\mathcal{P}_{\text{back}} \subset \mathcal{P}_{\text{DGP}}$ satisfying strong ignorability and SUTVA (see \Cref{eq:p_back}). This is one reason why CausalPFN is able to outperform Do-PFN in backdoor evaluation settings (see our benchmarking in \Cref{sec:benchmarking}, or \citep[Appendix~E]{balazadeh2026causalpfn}). On the other hand, CausalFM consists of three separate models trained on priors for different settings: one for the backdoor setting, one for instrumental variables, and one for the frontdoor setting \citep{ma2026causalfm}. Additionally, the authors prove that if a prior has non-identifiable support, the resulting PPD cannot recover the true causal effect even with infinite data, providing a strong argument for restricting to priors with identifiable support \citep[Theorem~4.3]{ma2026causalfm}. See \Cref{tab:first-cfms} for a comparison of these three CFMs and their priors.

\subsubsection{Transformer architectures for CFMs}

While a traditional causal estimator operates with training and test splits, CFMs operate with \emph{context} and \emph{query} data since no training is needed to perform inference. This is aligned with the current paradigm in TFMs \citep{hollmann2023tabpfn}; like their TFM cousins, Do-PFN, CausalPFN, and CausalFM are all transformer-based models (\Cref{tab:pfn-comparison}) that make predictions for novel queries via ICL while relying on a context set. CFMs receive a tabular observational dataset $\mathcal{D}_{\text{obs}} = \{(x_n, t_n, y_n)\}_{n = 1}^N$ as context, and a causal task as query. For example, a CATE query contains covariates $\{x_m\}_{m = 1}^M$, and the model's task is to estimate $\operatorname{CATE}(x_m)$. A CEPO query, on the other hand, would consist of covariates together with treatment values $\{(x_m, t_m)\}_{m = 1}^M$; the model's task is then to estimate $\mu_{t_m}(x_m)$.

Before being passed to the transformer, input data must be tokenized and embedded, with different CFMs performing different tokenization steps (e.g. row-based or row-and-feature-based). Notably, the embedding of the context dataset contains the observed covariates and treatments as well as the factual outcomes, while the embedding of the query does not contain any outcome information. The embedded representations are then passed to the transformer, which applies masked attention to the context and query tokens: each token attends to the context, and no token attends to the query. This mask, which is the same as that applied in TFMs, has two important consequences. First, the representation of the observational context data is independent of the queries. Second, query predictions depend only on the context data, not on each other. Batched query prediction is therefore independent of the content and order of the query set, once conditioned on the context $\mathcal{D}_{\text{obs}}$. 

The architecture of CFMs must take into account the distinguished role that the treatment variable plays. The simplest approach is to enforce that the treatment variable is the first column of the dataset, so that the model learns this in its internal representation. This is done by CausalPFN and Do-PFN, with Do-PFN also adding a treatment column indicator to the representation of each dataset. CausalFM, on the other hand, passes the treatment and covariates through different encoders before concatenating them and passing them to the transformer.

Inference proceeds by passing both the context and query to the model $\mathcal{M}_\theta$, which outputs its estimate $q_\theta$ of the desired causal estimand. For example, for CATE estimation:
\begin{equation}\label{eq:cate_model_ex}
    q_\theta(\text{CATE}(x_m) \mid x_m, \mathcal{D}_{\text{obs}}) = \mathcal{M}_\theta(\texttt{ctx}=\mathcal{D}_{\text{obs}}, \ \texttt{qry}=\{x_m\}, \ \texttt{est}=\texttt{CATE}).
\end{equation}
Since the CFM outputs the entire CATE-PPD, a point estimate of $\text{CATE}(x_m)$ can be obtained by taking the mean, $\hat\tau(x_m) \coloneqq \mathbb{E}[q_\theta(\text{CATE}(x_m) \mid x_m, \mathcal{D}_{\text{obs}})]$. In the example of CEPO estimation, we would have:
\begin{equation}\label{eq:cepo_model_ex}
    q_\theta(\mu_{t_m}(x_m) \mid x_m, t_m, \mathcal{D}_{\text{obs}}) = \mathcal{M}_\theta(\texttt{ctx}=\mathcal{D}_{\text{obs}}, \ \texttt{qry}=\{(x_m, t_m)\}, \ \texttt{est}=\texttt{CEPO}),
\end{equation}
and the point estimate $\hat\mu_{t_m}(x_m)\coloneqq \mathbb{E}[q_\theta(\mu_{t_m}(x_m) \mid x_m, t_m, \mathcal{D}_{\text{obs}})]$. The practical distinction between traditional and foundational causal models, enabled by the transformer architecture, is illustrated by the example algorithms in \Cref{fig:traditional-vs-cfm-algorithms}.

\begin{table}[t]
\centering
\small
\caption{Model size and transformer depth for different CFMs and TFMs.}
\label{tab:pfn-comparison}
\begin{tabular}{@{}lcc@{}}
\toprule
Model & \# Parameters & \# Transformer Layers  \\
\midrule
\multicolumn{3}{@{}l}{\emph{Tabular foundation models}} \\
TabICLv2 \citep{qu2026tabiclv2betterfasterscalable}
  & 29M & 12 \\
TabPFN-3 \citep{tabpfn_v3}
  & 58M & 24 \\
TabDPT-Turbo \citep{hosseinzadeh2026tabdptturbo}
  & 63M & 32  \\
\midrule
\multicolumn{3}{@{}l}{\emph{Causal foundation models}} \\
Do-PFN \citep{robertson2025do-pfn}
  & 7.3M & 12 \\
CausalPFN \citep{balazadeh2026causalpfn}
  & 20M & 20 \\
CausalFM \citep{ma2026causalfm}
  & 2.7M
  & 10 \\
\bottomrule
\end{tabular}
\vspace{10pt}
\end{table}

\begin{figure}
\begin{minipage}[t]{0.48\textwidth}
\textbf{Algorithm 1: Traditional causal estimator}
\begin{algorithmic}[1]
\Require Training set $\mathcal{D}_{\text{obs}} = \{(x_n, t_n, y_n\}_{n=1}^N$, query $\{x_m\}_{m=1}^M$, estimator class $\mathcal{M}_\theta$
\Ensure Estimate causal query $\hat g(x_m)$
\State Specify assumptions and nuisance components
\State Initialize parameters $\theta$
\While{not converged}
    \State Compute loss $\mathcal{L}(\theta; \mathcal{D}_{\text{obs}})$
    \State Update $\theta \leftarrow \theta - \eta \nabla_\theta \mathcal{L}$
\EndWhile
\State Estimate $\hat g(x_m) = \mathcal{M}_\theta(x_m)$
\State \Return $\hat g(x_m)$
\end{algorithmic}
\end{minipage}
\hfill
\vrule width 0.4pt
\hfill
\begin{minipage}[t]{0.48\textwidth}
\textbf{Algorithm 2: Causal foundation model}
\begin{algorithmic}[1]
\Require Context set $\mathcal{D}_{\text{obs}} = \{(x_n, t_n, y_n)\}_{n=1}^N$, query $\{x_m\}_{m = 1}^M$, pretrained CFM $\mathcal{M}_\theta$
\Ensure Estimate causal query $\hat g(x_m)$
\State Estimate $\hat g(x_m) = \mathcal{M}_\theta(\mathcal{D}_{\text{obs}}, x_m)$
\State \Return $\hat g(x_m)$
\end{algorithmic}
\end{minipage}
\captionof{figure}{A high-level comparison between traditional causal estimators and CFMs (shown with the example task of CATE estimation). The traditional estimators must be trained for each new task, updating model parameters on a training set, whereas the CFM workflow keeps pretrained parameters fixed and uses the training dataset as context.
}\label{fig:traditional-vs-cfm-algorithms}
\end{figure}

\subsection{Training CFMs}\label{subsec:training_cfms}

As mentioned above, CFMs are trained on synthetic data. In \Cref{subsec:pfns} we discussed that a core part of training PFN-style models is designing a highly diverse prior to ensure maximum coverage of possible tasks that could be encountered at inference time. SCM-based priors make this possible due to their ability to computationally represent DGPs and to generate both observational and interventional data efficiently (\Cref{subsec:scms}). We therefore consider a prior $\pi(\psi)$ over the space of possible SCMs, writing $S^\psi$ for an SCM corresponding to the DGP $\psi$.

The method of sampling from a prior and instantiating an SCM $S^\psi$ differs across models. Do-PFN's prior instantiates random DAGs directly using topological sorting, and then assigns structural equations generated as additive noise models \citep{robertson2025do-pfn}. CausalPFN uses random MLPs as in TabPFNv1 \citep{hollmann2023tabpfn}, subsampling a random set of nodes and using these to construct tabular datasets \citep{balazadeh2026causalpfn}. CausalFM samples clusters of DAGs using random MLPs, then uses Bayesian neural networks (BNNs) to assign values and sampling mechanisms for each cluster \citep{ma2026causalfm}.

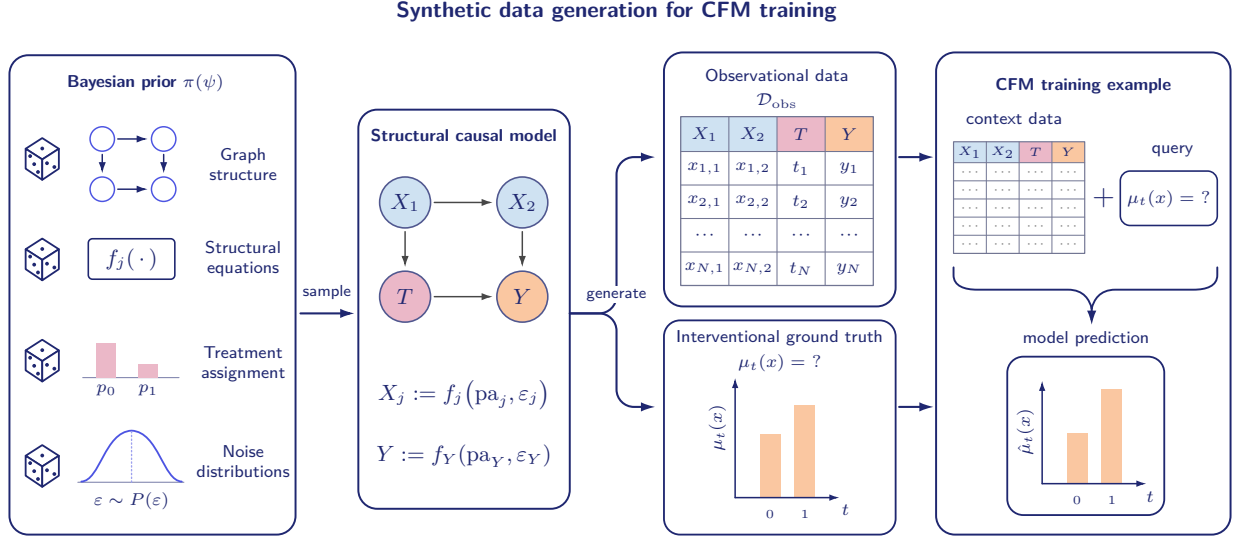
\begin{figure}[t]
    \centering
    \begingroup
    \definecolor{cfmnavy}{RGB}{31,39,112}
    \definecolor{cfmblue}{RGB}{76,84,226}
    \definecolor{cfmlightblue}{RGB}{207,226,242}
    \definecolor{cfmpink}{RGB}{235,184,200}
    \definecolor{cfmorange}{RGB}{250,199,161}
    \definecolor{cfmteal}{RGB}{45,123,139}
    \definecolor{cfmgrid}{RGB}{113,117,147}

    \resizebox{0.98\linewidth}{!}{%
    \begin{tikzpicture}[
        x=1cm,
        y=1cm,
        line cap=round,
        line join=round,
        every node/.style={font=\sffamily\scriptsize, text=cfmnavy},
        panel/.style={
            draw=cfmnavy,
            fill=white,
            line width=0.80pt,
            rounded corners=2.7mm
        },
        smallpanel/.style={
            draw=cfmnavy,
            fill=white,
            line width=0.75pt,
            rounded corners=2.3mm
        },
        title/.style={font=\sffamily\bfseries\small, text=cfmnavy},
        paneltitle/.style={font=\sffamily\bfseries\scriptsize, text=cfmnavy},
        flow/.style={
            -{Latex[length=2.1mm,width=1.35mm]},
            draw=cfmnavy,
            line width=0.85pt
        },
        causal/.style={
            -{Latex[length=1.65mm,width=1.05mm]},
            draw=black!72,
            line width=0.60pt,
            shorten <=1.2pt,
            shorten >=1.2pt
        },
        scmnode/.style={
            circle,
            draw=cfmnavy!78,
            line width=0.65pt,
            minimum size=7.3mm,
            inner sep=0pt,
            font=\rmfamily\small,
            text=cfmnavy
        },
        chartline/.style={draw=cfmblue, line width=0.75pt},
        axis/.style={-{Latex[length=1.5mm,width=0.95mm]}, draw=cfmnavy, line width=0.62pt},
        chartpoint/.style={circle, draw=cfmblue, fill=white, line width=0.55pt, minimum size=2.5pt, inner sep=0pt}
    ]

        \node[title] at (8.85,8.05) {Synthetic data generation for CFM training};

        \draw[panel] (0.15,0.55) rectangle (4.25,7.42);
        \node[paneltitle] at (2.10,7.03) {Bayesian prior $\pi(\psi)$};

        \def\dicelinewidth{0.75pt}
        \def\dicepipradius{0.080}
        \foreach \yy in {5.92,4.48,3.03,1.52}{
            \begin{scope}[shift={(0.62,\yy)},scale=0.34]
                \coordinate (top) at (0,0.92);
                \coordinate (left) at (-0.70,0.47);
                \coordinate (mid) at (0,0.02);
                \coordinate (right) at (0.70,0.47);
                \coordinate (bottom) at (0,-0.84);
                \draw[draw=cfmnavy,line width=\dicelinewidth,fill=white] (top)--(left)--(mid)--(right)--cycle;
                \draw[draw=cfmnavy,line width=\dicelinewidth,fill=white] (left)--(mid)--(bottom)--(-0.70,-0.37)--cycle;
                \draw[draw=cfmnavy,line width=\dicelinewidth,fill=white] (mid)--(right)--(0.70,-0.37)--(bottom)--cycle;
                \fill[cfmnavy] (0,0.48) circle (\dicepipradius);
                \fill[cfmnavy] (-0.51,0.16) circle (\dicepipradius);
                \fill[cfmnavy] (-0.34,-0.18) circle (\dicepipradius);
                \fill[cfmnavy] (-0.18,-0.51) circle (\dicepipradius);
                \fill[cfmnavy] (0.20,-0.14) circle (\dicepipradius);
                \fill[cfmnavy] (0.51,-0.3) circle (\dicepipradius);
            \end{scope}
        }

        \def\textcenterpoint{3.5}
        \foreach \p in {(1.48,6.22),(2.36,6.22),(1.48,5.50),(2.36,5.50)}
            \node[circle,draw=cfmblue,fill=white,line width=0.60pt,minimum size=3.5mm,inner sep=0pt] at \p {};
        \draw[causal,draw=cfmnavy] (1.66,6.22)--(2.18,6.22);
        \draw[causal,draw=cfmnavy] (1.48,6.04)--(1.48,5.68);
        \draw[causal,draw=cfmnavy] (1.66,5.50)--(2.18,5.50);
        \draw[causal,draw=cfmnavy] (2.36,6.04)--(2.36,5.68);
        \node[align=center] at (\textcenterpoint,5.86) {Graph\\structure};

        \draw[draw=cfmnavy,line width=0.65pt,rounded corners=0.6mm] (1.28,4.21) rectangle (2.52,4.79);
        \node[font=\rmfamily\small,text=cfmnavy] at (1.90,4.50) {$f_j(\,\cdot\,)$};
        \node[align=center] at (\textcenterpoint,4.50) {Structural\\equations};

        \draw[cfmnavy!55,line width=0.45pt] (1.20,2.80)--(2.58,2.80);
        \fill[cfmpink] (1.40,2.80) rectangle (1.68,3.29);
        \fill[cfmpink] (2.00,2.80) rectangle (2.28,3);
        \node[font=\rmfamily\scriptsize] at (1.54,2.62) {$p_0$};
        \node[font=\rmfamily\scriptsize] at (2.14,2.62) {$p_1$};
        \node[align=center] at (\textcenterpoint,3.00) {Treatment\\assignment};

        \draw[cfmnavy!55,line width=0.45pt] (1.13,1.30)--(2.65,1.30);
        \draw[cfmblue,line width=0.75pt]
            (1.18,1.31) .. controls (1.45,1.32) and (1.51,2.02) .. (1.90,2.04)
            .. controls (2.29,2.02) and (2.35,1.32) .. (2.62,1.31);
        \draw[cfmblue!55,densely dotted,line width=0.55pt] (1.90,1.30)--(1.90,2.04);
        \node[font=\rmfamily\scriptsize,text=cfmnavy] at (1.90,1.03) {$\varepsilon\sim P(\varepsilon)$};
        \node[align=center] at (\textcenterpoint,1.60) {Noise\\distributions};

        \draw[panel] (5.15,0.92) rectangle (8.18,6.64);
        \node[paneltitle] at (6.665,6.28) {Structural causal model};

        \node[scmnode,fill=cfmlightblue] (xone) at (5.82,5.31) {$X_1$};
        \node[scmnode,fill=cfmlightblue] (xtwo) at (7.50,5.31) {$X_2$};
        \node[scmnode,fill=cfmpink] (trt) at (5.82,3.96) {$T$};
        \node[scmnode,fill=cfmorange] (out) at (7.50,3.96) {$Y$};
        \draw[causal] (xone)--(xtwo);
        \draw[causal] (xone)--(trt);
        \draw[causal] (xtwo)--(out);
        \draw[causal] (trt)--(out);

        \node[font=\rmfamily\small,text=cfmnavy] at (6.665,2.56)
            {$X_j := f_j\!\left(\mathrm{pa}_j,\varepsilon_j\right)$};
        \node[font=\rmfamily\small,text=cfmnavy] at (6.665,1.68)
            {$Y := f_Y\!\left(\mathrm{pa}_Y,\varepsilon_Y\right)$};

        \draw[flow] (4.32,3.72)--(5.1,3.72);
        \node[font=\sffamily\fontsize{6.5pt}{7.5pt}\selectfont] at (4.70,4.03) {sample};

        \begin{scope}[xshift=0.25cm]
        \draw[panel] (9.28,3.82) rectangle (12.60,7.46);
        \node[font=\sffamily\scriptsize] at (10.90,7.14) {Observational data};
        \node[font=\rmfamily\scriptsize] at (10.90,6.77) {$\mathcal{D}_{\mathrm{obs}}$};

        \fill[cfmlightblue] (9.52,6.04) rectangle (10.90,6.54);
        \fill[cfmpink]      (10.90,6.04) rectangle (11.59,6.54);
        \fill[cfmorange]    (11.59,6.04) rectangle (12.28,6.54);
        \draw[draw=cfmgrid,line width=0.52pt] (9.52,4.12) rectangle (12.28,6.54);
        \foreach \xx in {10.21,10.90,11.59}
            \draw[draw=cfmgrid,line width=0.43pt] (\xx,4.12)--(\xx,6.54);
        \foreach \yy in {4.60,5.08,5.56,6.04}
            \draw[draw=cfmgrid,line width=0.43pt] (9.52,\yy)--(12.28,\yy);
        \node[font=\rmfamily\scriptsize] at (9.865,6.29) {$X_1$};
        \node[font=\rmfamily\scriptsize] at (10.555,6.29) {$X_2$};
        \node[font=\rmfamily\scriptsize] at (11.245,6.29) {$T$};
        \node[font=\rmfamily\scriptsize] at (11.935,6.29) {$Y$};
        \node[font=\rmfamily\scriptsize] at (9.865,5.80) {$x_{1,1}$};
        \node[font=\rmfamily\scriptsize] at (10.555,5.80) {$x_{1,2}$};
        \node[font=\rmfamily\scriptsize] at (11.245,5.80) {$t_1$};
        \node[font=\rmfamily\scriptsize] at (11.935,5.80) {$y_1$};
        \node[font=\rmfamily\scriptsize] at (9.865,5.32) {$x_{2,1}$};
        \node[font=\rmfamily\scriptsize] at (10.555,5.32) {$x_{2,2}$};
        \node[font=\rmfamily\scriptsize] at (11.245,5.32) {$t_2$};
        \node[font=\rmfamily\scriptsize] at (11.935,5.32) {$y_2$};
        \foreach \xx in {9.865,10.555,11.245,11.935}
            \node[font=\rmfamily\scriptsize] at (\xx,4.84) {$\cdotp\mkern-2mu\cdotp\mkern-2mu\cdotp$};
        \node[font=\rmfamily\scriptsize] at (9.865,4.36) {$x_{N,1}$};
        \node[font=\rmfamily\scriptsize] at (10.555,4.36) {$x_{N,2}$};
        \node[font=\rmfamily\scriptsize] at (11.245,4.36) {$t_N$};
        \node[font=\rmfamily\scriptsize] at (11.935,4.36) {$y_N$};

        \draw[smallpanel] (9.28,0.55) rectangle (12.60,3.62);
        \node[font=\sffamily\scriptsize,text=cfmnavy] at (10.94,3.37) {Interventional ground truth};
        \node[font=\rmfamily\scriptsize,text=cfmnavy] at (10.94,3.01) {$\mu_t(x)=\ ?$};

        \begin{scope}[xshift=0.5cm]
        \node[font=\rmfamily\scriptsize,rotate=90] at (9.55,2.08) {$\mu_t(x)$};
        \node[font=\rmfamily\scriptsize] at (11.38,0.90) {$t$};
        
        \fill[cfmorange] (10.15,1.08) rectangle (10.45,1.99);
        \node[font=\rmfamily\tiny] at (10.30,0.84) {$0$};
        
        \fill[cfmorange] (10.65,1.08) rectangle (10.95,2.4);
        \node[font=\rmfamily\tiny] at (10.80,0.84) {$1$};

        \draw[axis] (9.80,1.08)--(11.30,1.08);
        \draw[axis] (9.80,1.08)--(9.80,2.80);
        \end{scope}
        \end{scope}

        \draw[flow, rounded corners=2.5mm] (8.18,3.72) -| (8.85,5.96) -- (9.41,5.96);
        \draw[flow, rounded corners=2.5mm] (8.18,3.72) -| (8.85,2.38) -- (9.41,2.38);
        \node[font=\sffamily\fontsize{6.5pt}{7.5pt}\selectfont,fill=white,inner sep=0.8pt] at (8.85,3.99) {generate};
        
        \draw[panel] (13.43,0.55) rectangle (17.65,7.46);
        \node[paneltitle] at (15.54,6.99) {CFM training example};

        \node[font=\sffamily\scriptsize] at (14.55,6.52) {context data};
        \node[font=\sffamily\scriptsize] at (16.82,6.05) {query};

        \fill[cfmlightblue] (13.68,5.88) rectangle (14.62,6.18);
        \fill[cfmpink]      (14.62,5.88) rectangle (15.09,6.18);
        \fill[cfmorange]    (15.09,5.88) rectangle (15.56,6.18);
        \draw[draw=cfmgrid,line width=0.48pt] (13.68,4.58) rectangle (15.56,6.18);
        \foreach \xx in {14.15,14.62,15.09}
            \draw[draw=cfmgrid,line width=0.40pt] (\xx,4.58)--(\xx,6.18);
        \foreach \yy in {4.84,5.10,5.36,5.62,5.88}
            \draw[draw=cfmgrid,line width=0.40pt] (13.68,\yy)--(15.56,\yy);
        \node[font=\rmfamily\tiny] at (13.915,6.03) {$X_1$};
        \node[font=\rmfamily\tiny] at (14.385,6.03) {$X_2$};
        \node[font=\rmfamily\tiny] at (14.855,6.03) {$T$};
        \node[font=\rmfamily\tiny] at (15.325,6.03) {$Y$};
        \foreach \yy in {5.75,5.49,5.23,4.97,4.71}{
            \foreach \xx in {13.915,14.385,14.855,15.325}
                \node[font=\rmfamily\tiny,text=black!55] at (\xx,\yy) {$\cdotp\mkern-2mu\cdotp\mkern-2mu\cdotp$};
        }

        \node[font=\sffamily\large] at (15.83,5.38) {$+$};
        \draw[draw=cfmnavy,line width=0.65pt,rounded corners=1.4mm] (16.07,4.98) rectangle (17.46,5.76);
        \node[font=\rmfamily\scriptsize,text=cfmnavy] at (16.765,5.37) {$\mu_t(x)=\ ?$};

        \draw[draw=cfmnavy,line width=0.78pt]
            (13.68,4.38)--(13.68,4.27)
            .. controls (13.68,4.11) and (13.84,4.06) .. (14.02,4.06)
            --(15.32,4.06)
            .. controls (15.55,4.06) and (15.63,3.91) .. (15.66,3.79);
        \draw[draw=cfmnavy,line width=0.78pt]
            (17.46,4.38)--(17.46,4.27)
            .. controls (17.46,4.11) and (17.30,4.06) .. (17.12,4.06)
            --(16.00,4.06)
            .. controls (15.77,4.06) and (15.69,3.91) .. (15.66,3.79);
        \draw[flow] (15.66,3.82)--(15.66,3.55);

        \begin{scope}[xshift=0.4cm]
        \node[font=\sffamily\scriptsize,text=cfmnavy] at (15.18,3.36) {model prediction};
        \draw[smallpanel] (14.05,0.83) rectangle (16.30,3.10);
        \node[font=\rmfamily\scriptsize,rotate=90] at (14.31,2.03) {$\hat\mu_t(x)$};
        \node[font=\rmfamily\scriptsize] at (16.10,1.10) {$t$};
        
        \fill[cfmorange] (14.90,1.28) rectangle (15.20,2.01);
        \node[font=\rmfamily\tiny] at (15.05,1.04) {$0$};
        
        \fill[cfmorange] (15.40,1.28) rectangle (15.70,2.63);
        \node[font=\rmfamily\tiny] at (15.55,1.04) {$1$};

        \draw[axis] (14.55,1.28)--(16.05,1.28);
        \draw[axis] (14.55,1.28)--(14.55,2.76);
        \end{scope}

        \draw[flow] (12.9,5.96)--(13.4,5.96);
        \draw[flow] (12.9,2.38)--(13.4,2.38);
        
    \end{tikzpicture}%
    }
    \endgroup
    \caption{Sampling data from a synthetic causal prior. \emph{Step 1.} Sample an SCM from $\pi$. \emph{Step 2.} Generate an observational dataset $\mathcal{D}_{\mathrm{obs}}$ from the SCM. \emph{Step 3.} Simulate interventional supervised learning targets, such as the CEPO or CATE. \emph{Step 4.} Provide the observational data to the CFM as context and train it to estimate the target.}
    \label{fig:data_from_scm}
\end{figure}

At a high level, a CFM training run has the following form (see \Cref{fig:data_from_scm} for a visual overview):
\begin{enumerate}
    \item Sample an SCM $S^\psi \sim \pi$. In practice, this might mean initializing the weights of an MLP; selecting which edges to prune and which nodes to select as covariates $X$, treatment $T$, and outcome $Y$; and deciding on the probability distributions $P_{U}$ from which noise variables $U_k$ will be sampled.
    \item Generate an observational dataset $\mathcal{D}_{\text{obs}} \sim P^\psi_{\text{obs}}$ of size $N$ from $S^\psi$. This is done by sampling the noise variables $U_k \sim P_U$ and using the structural equations of $S^\psi$ to compute the values of $X$, $T$, and $Y$.
    \item Simulate interventions to create an interventional dataset $\mathcal{D}_{\text{int}} \sim P^\psi$ of size $M$ from $S^\psi$. This is done differently for different interventional targets (e.g. CEPO-PPD, CID-PPD).
    \item Perform supervised learning by asking the model to predict the true causal target from $\mathcal{D}_{\text{obs}}$, using a causal prior-data loss (e.g. \Cref{eq:cepo_data_prior_loss}).
\end{enumerate}

Let us focus more on step 3, which is the heart of causal estimation tasks. If this were a predictive PFN training loop, step 3 would generate data from the same distribution $P^\psi_{\text{obs}}$ as step 2, asking the model to predict data from the same distribution that it is shown as context. In causal inference, we are asking the model to estimate causal quantities, like the CEPO and the CID, from observational data alone. This requires generating counterfactual, or interventional, data. 

\begin{figure}[t]
\centering
\begin{tikzpicture}[
    x=1cm,
    y=1cm,
    line cap=round,
    line join=round,
    every node/.style={font=\small},
    scmnode/.style={
        circle,
        draw=black!70,
        fill=white,
        line width=0.75pt,
        minimum size=8.5mm,
        inner sep=0pt
    },
    noise/.style={
        circle,
        draw=black!55,
        fill=gray!3,
        dashed,
        line width=0.65pt,
        minimum size=8.5mm,
        inner sep=1pt,
        font=\scriptsize
    },
    intervention/.style={
        draw=orange!85!black,
        fill=orange!15,
        rounded corners=2pt,
        line width=0.9pt,
        minimum width=13mm,
        minimum height=8.5mm,
        inner sep=1pt
    },
    source/.style={
        draw=black!60,
        fill=gray!6,
        rounded corners=2.5pt,
        line width=0.65pt,
        align=center,
        inner xsep=7pt,
        inner ysep=5pt, 
        font=\small
    },
    dataset/.style={
        draw=teal!60!black,
        fill=teal!5,
        rounded corners=2.5pt,
        line width=0.65pt,
        minimum width=5.45cm,
        align=center,
        inner xsep=5pt,
        inner ysep=4pt
    },
    interventional_dataset/.style={
        draw=blue!60!black,
        fill=blue!5,
        rounded corners=2.5pt,
        line width=0.65pt,
        minimum width=5.45cm,
        align=center,
        inner xsep=5pt,
        inner ysep=4pt
    },
    graphbox/.style={
        draw=black!18,
        rounded corners=4pt,
        line width=0.5pt,
        inner sep=2.5mm
    },
    causal/.style={
        -{Latex[length=2.0mm,width=1.35mm]},
        draw=black!75,
        line width=0.75pt,
        shorten <=1.2pt,
        shorten >=1.2pt
    },
    process/.style={
        -{Latex[length=2.2mm,width=1.5mm]},
        draw=black!65,
        line width=0.85pt,
        rounded corners=2pt
    },
    connector/.style={
        draw=black!65,
        line width=0.85pt
    },
    cutedge/.style={
        draw=orange!75!black,
        densely dashed,
        line width=0.7pt,
        shorten <=2pt,
        shorten >=2pt
    },
    cutmark/.style={
        text=orange!85!black,
        fill=white,
        font=\scriptsize\bfseries,
        inner sep=0.5pt
    }
]
    \node[source] (episode) at (0, 4.60) {Sampled SCM $S^\psi \sim \pi$\\[-0mm]
        and noises $\{(u_{n, 1}^{\text{obs}},u_{n, 2}^{\text{obs}},u_{n, 3}^{\text{obs}})\}_{n=1}^{N}$, \quad
        $\{(u_{m, 1}^{\text{int}},u_{m, 2}^{\text{int}},u_{m, 3}^{\text{int}})\}_{m=1}^{M}$};

    \coordinate (fork) at (0, 3.72);
    \coordinate (obs-entry) at (-4.0, 3.40);
    \coordinate (int-entry) at (4.0, 3.40);
    \draw[connector] (episode.south) -- (fork);
    \draw[process] (fork) -| (obs-entry);
    \draw[process] (fork) -| (int-entry);

    \node[font=\small\bfseries] (obs-title) at (-4.0, 3.08) {(a) Observational pass};
    \node[font=\small\bfseries] (int-title) at (4.0, 3.08) {(b) Interventional pass};
    \draw[gray!45, densely dashed, line width=0.55pt] (0, 3.40) -- (0, -1.35);

    \node[scmnode] (x-obs) at (-4.0, 2.05) {$X$};
    \node[scmnode] (t-obs) at (-5.15, 0.85) {$T$};
    \node[scmnode] (y-obs) at (-2.85, 0.85) {$Y$};
    \node[noise] (ux-obs) at (-2.50, 2.05) {$U_1$};
    \node[noise] (ut-obs) at (-6.45, 0.85) {$U_2$};
    \node[noise] (uy-obs) at (-1.55, 0.85) {$U_3$};

    \draw[causal] (x-obs) -- (t-obs);
    \draw[causal] (x-obs) -- (y-obs);
    \draw[causal] (t-obs) -- (y-obs);
    \draw[causal] (ux-obs) -- (x-obs);
    \draw[causal] (ut-obs) -- (t-obs);
    \draw[causal] (uy-obs) -- (y-obs);

    \node[graphbox, fit=(x-obs)(t-obs)(y-obs)(ux-obs)(ut-obs)(uy-obs)] (g-obs) {};
    \node[anchor=north west, font=\scriptsize, text=black!30, inner sep=1.5pt]
        at (g-obs.north west) {$S^\psi$};
    \node[dataset] (d-obs) at (-4.0, -0.85) {$\mathcal{D}_{\mathrm{obs}}
        = \{(x_n,t_n,y_n)\}_{n=1}^{N}$\\[-0.5mm]
        \scriptsize\color{teal!60!black} context data};
    \draw[process] (g-obs.south) -- (d-obs.north);

    \node[scmnode] (x-int) at (4.0, 2.05) {$X$};
    \node[intervention] (t-int) at (2.85, 0.85) {$T=t^*$};
    \node[scmnode] (y-int) at (5.15, 0.85) {$Y$};
    \node[noise] (ux-int) at (5.50, 2.05) {$U_1$};
    \node[noise, draw=gray!40, text=gray!50, fill=gray!2] (ut-int) at (1.55, 0.85) {$U_2$};
    \node[noise] (uy-int) at (6.45, 0.85) {$U_3$};

    \draw[causal] (x-int) -- (y-int);
    \draw[causal] (t-int) -- (y-int);
    \draw[causal] (ux-int) -- (x-int);
    \draw[causal] (uy-int) -- (y-int);

    \draw[cutedge] (x-int) -- (t-int);
    \draw[cutedge] (ut-int) -- (t-int);
    \path (x-int) -- (t-int) coordinate[pos=0.58] (cut-x);
    \path (ut-int) -- (t-int) coordinate[pos=0.58] (cut-u);
    \node[cutmark] at (cut-x) {$\times$};
    \node[cutmark] at (cut-u) {$\times$};

    \node[graphbox, fit=(x-int)(t-int)(y-int)(ux-int)(ut-int)(uy-int)] (g-int) {};
    \node[anchor=north west, font=\scriptsize, text=black!30, inner sep=1.5pt]
        at (g-int.north west) {$S^\psi$};
    \node[align=center, text=orange!80!black, fill=white, inner sep=1pt, font=\scriptsize] at (2.85, 0.08)
        {$\mathrm{do}(T=t^*)$\\replaces $f_T$};

    \node[interventional_dataset] (d-int) at (4.0, -0.85) {$\mathcal{D}_{\mathrm{int}}
        = \{(x_m,t^*,y_m(t^*))\}_{m=1}^{M}$\\[-0.5mm]
        \scriptsize\color{blue!60!black} interventional targets};
    \draw[process] (g-int.south) -- (d-int.north);
\end{tikzpicture}
\caption{Observational and interventional dataset generation from an SCM $S^\psi$. \captiona~The full set of original structural equations map a batch of exogenous noise vectors $\{(u_{n, 1}^{\text{obs}},u_{n, 2}^{\text{obs}},u_{n, 3}^{\text{obs}})\}_{n=1}^{N}$ to observational data $\mathcal{D}_{\text{obs}}$. \captionb~The intervention $\operatorname{do}(T = t^*)$ sets the value of treatment for the $m$th query individual to be $t^*$, replacing the structural equation $f_T$ of $T$ with $T=t^*$. Using the noise samples $\{(u_{m, 1}^{\text{int}},u_{m, 2}^{\text{int}},u_{m, 3}^{\text{int}})\}_{m=1}^{M}$, the remaining structural equations then produce an interventional dataset $\mathcal{D}_{\text{int}}$ with ground-truth potential outcomes $y_m(t^*)$.}
\label{fig:data_from_scm_detailed}
\end{figure}
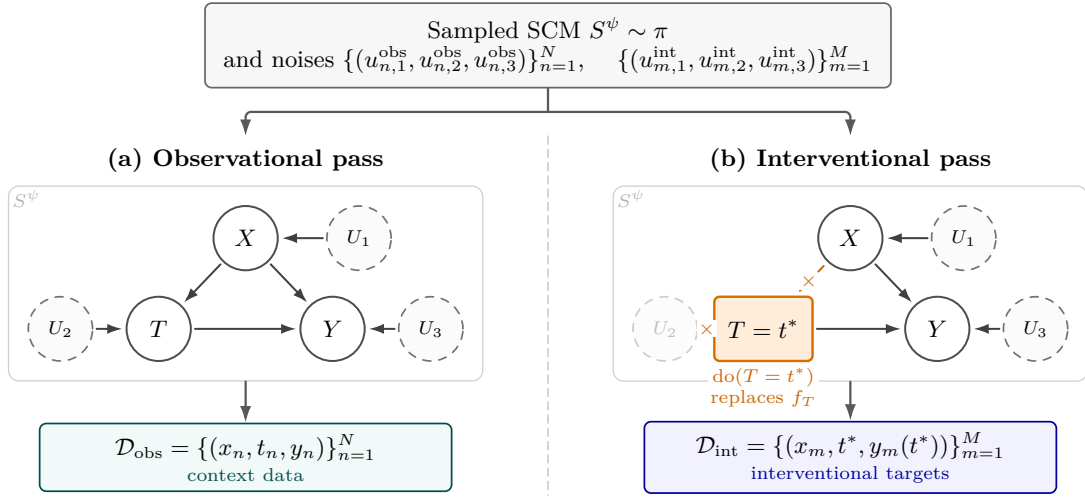

To generate potential outcomes $y(t)$, which is required by the CID-PPD causal prior-data loss (\Cref{eq:counterfactual_data_prior_loss}), we must generate data from the conditional interventional distribution $P^\psi(\cdot \mid \operatorname{do}(T = t), X = x)$. As we saw in \Cref{subsec:scms}, we mechanically replace the $T$ structural equation by $T = t^*$ with a fixed ``query'' treatment $t^* \in \mathcal{T}$. Graphically, this corresponds to removing the incoming edges to the treatment node $T$. We can then sample $U_k \sim P_U$ and use the remaining structural equations to generate an interventional sample $(x, t^*, y(t^*))$, representing the outcome of an individual with covariates $x$ as a result of receiving treatment $t^*$. This is depicted in \Cref{fig:data_from_scm_detailed}.

To generate ground-truth CEPO labels, we can take an expectation over the exogenous noise variables $U_k$:
\begin{equation}\label{eq:scm_cepo}
    \mu_t(x) = \mathbb{E}_{U_k \sim P_{U \mid X}}[Y(t) \mid X = x].
\end{equation}
For example, for the SCM depicted in \Cref{fig:data_from_scm_detailed}, we have
\begin{equation}
    \mu_t(x) = \mathbb{E}_{U_3 \sim P_U}[Y(t) \mid X = x].
\end{equation}
The generated SCMs must be diverse and complex, and hence \Cref{eq:scm_cepo} cannot generally be written in closed form. Although it is straightforward to approximate using Monte Carlo sampling, obtaining an accurate estimate of \cref{eq:scm_cepo} may require many noise samples and passes through the SCM. This must be repeated for each $(x, t)$ pair for which $\mu_t(x)$ is to be calculated, severely slowing down training time. An alternative employed by \citet{balazadeh2026causalpfn} is to replace the node $Y$ in the SCM with a CEPO node $\mu$, so that a forward pass through the SCM generates $\mu_t(x)$ directly instead of $Y(t)$. In order to capture realistic fluctuations in the true outcome $Y$, a new node $\xi$ is introduced to the SCM which functions as a noise node. After centering and scaling $\xi$, the potential outcome $Y(t)$ is computed as $Y(t) = \mu_t(x) + \xi(x, t)$. With this method, both the CEPO and the potential outcome are computed by the SCM in one forward pass.

So far, no research has been published on the quality of priors for CFMs. However, it seems plausible that the diversity aspect of prior design should align with research on priors for TFMs. \citet{zhang2025mitramixedsyntheticpriors} quantify prior diversity and support its importance for TFM performance. Recently, \citet{turkmen2026evaluatingdatapriorstabular} have created an interface to compare different TFM priors and their relative strengths and weaknesses on downstream tasks. However, similar work has not been undertaken yet for causal priors.

In summary, current transformer-based CFMs are designed to predict a causal quantity and are pretrained with the corresponding causal prior-data loss. Each iteration of training involves sampling an SCM from a prior, using it to generate a synthetic observational dataset, and also simulating corresponding interventional data. The CFM outputs an approximation of the causal PPD, using the observational data as context and unlabeled data as the query in a transformer architecture, with special care taken to tokenize and embed covariates, treatments, and outcomes. Based on the outputs, the CFM is trained to minimize the negative log-likelihood it assigns to the true sample from the causal PPD, which derives from the simulated interventional data. Finally, at inference the same context-query setup is used to directly predict causal quantities given only observational data, performing causal inference on unseen data in a single forward pass. All CFMs share this basic function, with important design choices including which PPD to model, the specification of the SCM prior including identifiability, and architectural considerations like input embeddings and attention mechanisms.

\section{Benchmarking CFMs}\label{sec:benchmarking}
In this section, we evaluate emerging CFMs against established causal machine learning estimators. While \Cref{sec:cfms-core} categorized the architectural paradigms of Do-PFN \citep{robertson2025do-pfn}, CausalPFN \citep{balazadeh2026causalpfn}, and CausalFM \citep{ma2026causalfm}, 
existing empirical comparisons in the literature remain largely limited to synthetic structural simulations under varying data-generation setups. 

The main exception is the Jobs experiment of \citet[Section~6.1]{ma2026causalfm}, which compares all three models on a semi-synthetic version of the Jobs dataset \citep{smith2005does}, derived from the Lalonde study \citep{lalonde1986evaluating}; there, outcomes are simulated by the authors to allow evaluation against ground truth, and only conditional effect error is reported.

One contribution of our work is establishing a standardized, fair empirical comparison across these three CFMs for causal inference on more complex semi-synthetic observational data. Since purely observational datasets lack counterfactual data, and purely synthetic simulations often oversimplify real-world mechanisms, we evaluate models on the semi-synthetic \textbf{RealCause-Lalonde} benchmark \citep{lalonde1986evaluating, neal2020realcause}. RealCause has been widely adopted across the causal machine 
learning community as a realistic evaluation standard  \citep[e.g.,][]{mahajan2024empirical, manela2024marginal, vanderlaan2026doublyrobustinferencecalibration}, 
as well as in evaluations of CFMs
\citep[e.g.,][]{robertson2025do-pfn, balazadeh2026causalpfn}.
This allows us to evaluate both conditional effect heterogeneity ($\mathrm{CATE}$) and population-level effects ($\mathrm{ATE}$) against known ground truth under strong confounding, creating an evaluation regime that effectively differentiates performance across models.

\subsection{Experimental Setup}\label{subsec:exp-setup}

\paragraph{Benchmark Data.} We evaluate on the semi-synthetic Lalonde-CPS (16,177 samples) and Lalonde-PSID (2,675 samples) cohorts \citep{lalonde1986evaluating, neal2020realcause}. RealCause fits generative models to treated individuals and non-experimental control individuals. By generating simulated outcomes as explicit functions of observed covariates alone, the benchmark guarantees conditional ignorability ($Y(1), Y(0) \perp\!\!\perp T \mid X$) by construction while preserving the complex, empirical covariate and selection distributions. To account for sampling variability in the generative process, all metrics are aggregated across 10 independent semi-synthetic realizations and reported as mean $\pm$ standard error.

\paragraph{Baselines.} We compare CFMs against standard causal machine learning estimators. Classical baselines include meta-learners (S-Learner, T-Learner, X-Learner \citep{kunzel2019metalearners}), propensity and doubly robust methods (IPW, DR-Learner \citep{robins1994estimation, kennedy2023towards}), and Double Machine Learning (Debiased ML, \citep{chernozhukov2018double}). All estimators except IPW are implemented and tuned via EconML \citep{battocchi2019econml}. To provide strong, competitive baselines, all underlying nuisance models are selected via FLAML AutoML (v2.3.5; \citet{wang2021flamlfastlightweightautoml}) run independently per nuisance, per estimator, per RealCause realization, and per cohort with a 900-second budget and 3-fold cross-validation.  For outcome regressions, FLAML optimizes $R^2$ across candidate families including LightGBM, depth-limited and unlimited XGBoost, Random Forest, Extra Trees, and $k$-NN. For propensity models (required by X-Learner, DML, DR-Learner, and IPW), FLAML optimizes ROC-AUC across classifier versions of these families as well as $\ell_1$- and $\ell_2$-regularized logistic regression.  For the final stages, DML uses an unregularized linear model and DR-Learner uses a causal forest (\texttt{ForestDRLearner} with 1,000 trees, 5-fold cross-validation, and degree-3 polynomial features). This hyperparameter regime produces highly optimized versions of the classical baselines.

For CFMs, we evaluate frozen, pre-trained checkpoints of Do-PFN, CausalPFN, and CausalFM in the backdoor setting. These models are downloaded out-of-the-box from their respective sources and applied without any parameter or hyperparameter tuning. We note that a fourth model designed by \citet{dhir2025estimatinginterventionaldistributions} could be considered a CFM, but no pre-trained weights are available to make a direct comparison here.

\paragraph{Metrics.} 

Consider $N$ evaluation individuals with covariates and ground-truth CATE values $\{(x_n, \tau(x_n))\}_{n=1}^N$. Let $\lambda = \frac{1}{N}\sum_{n=1}^N \tau(x_n)$ denote the true ATE. We write $\hat\tau(x_n)$ to denote the predicted CATE values and $\hat\lambda$ to denote the predicted ATE. We evaluate the following metrics:
\begin{equation}\label{eq:metrics}
    \operatorname{PEHE}(\hat{\tau}) = \sqrt{\frac{1}{N} \sum_{n=1}^N \big(\tau(x_n) - \hat{\tau}(x_n)\big)^2}, \qquad \operatorname{RelativeError}(\hat{\lambda}) = \frac{|\hat{\lambda} - \lambda|}{|\lambda|}.
\end{equation}
Precision in estimation of heterogeneous effects ($\operatorname{PEHE}$) measures root mean squared error in conditional effect estimation across individuals, while the relative error quantifies relative population-level bias via the ATE. 

For Do-PFN which approximates the CID-PPD, point estimates for the potential outcomes $\mu_0(x)$ and $\mu_1(x)$ are first computed by taking the mean of the model's approximation $q_\theta$ to the CID-PPD (see \cref{eq:cepo_interventional}):
\begin{equation}
    \hat\mu_0(x) = \mathbb{E}[q_\theta(y \mid x, t = 0, \mathcal{D}_{\text{obs}})], \qquad \hat\mu_1(x) = \mathbb{E}[q_\theta(y \mid x, t = 1,  \mathcal{D}_{\text{obs}})].
\end{equation}
The CATE estimate is then returned as $\hat\tau(x) = \hat\mu_1(x) - \hat\mu_0(x)$ (see \cref{eq:cate_def}).

For CausalPFN which approximates the CEPO-PPD, point estimates for the potential outcomes are first obtained by returning the means of the approximated CEPO-PPDs:
\begin{equation}
    \hat\mu_0(x) = \mathbb{E}[q_\theta(\mu_{t}(x) \mid x, t=0, \mathcal{D}_{\text{obs}})], \qquad \hat\mu_1(x) = \mathbb{E}[q_\theta(\mu_{t}(x) \mid x, t=1, \mathcal{D}_{\text{obs}})].
\end{equation}
As above, the CATE estimate is then returned as $\hat\tau(x) = \hat\mu_1(x) - \hat\mu_0(x)$.

CausalFM's CATE model is trained to estimate the CDTE-PPD. It does so using a Gaussian mixture model (GMM) with 5 components, 
\begin{equation}
    q_\theta(Y(1) - Y(0) \mid X = x, \mathcal{D}_{\text{obs}}) = \sum_{k = 1}^5 w_k \mathcal{N}\big(Y(1) - Y(0) \mid \mu_k(x), \sigma_k^2(x)\big).
\end{equation}
The CATE estimate is then $\hat \tau(x) = \sum_{k = 1}^5 w_k \mu_k(x)$.

We additionally report wall-clock CPU runtime per cohort for going from observational data to predictions. This means we include training, tuning, and inference for classical baselines, but only inference for CFMs since this is the only step we perform. Hence, our measurements reflect actual practitioner wall-clock time needed to apply each given method in practice. 

Finally, we report the average rank among the models tested. For each metric, methods are ranked against each other within a single realization, and these ranks are then averaged over the realizations of both Lalonde-CPS and Lalonde-PSID cohorts, again reported as $\text{mean} \pm \text{standard error}$. Since IPW does not produce individual-level estimates, PEHE ranks are computed among the remaining methods.

\paragraph{Reproducibility \& Code Artifacts.} The full pipeline for loading the RealCause-Lalonde benchmark, aggregating empirical results, and generating comparison tables and figures is provided as an interactive Jupyter notebook at \jupyter{https://github.com/layer6ai-labs/cfms/blob/main/notebooks/Lalonde_benchmark_results.ipynb}. 
Since running 10 seeds of the benchmark takes a non-trivial time and access to a GPU, we also release our full dataset of the results at \href{https://github.com/layer6ai-labs/cfms/tree/main/data}{\texttt{github.com/layer6ai-labs/cfms/tree/main/data}}.

\subsection{Results and Discussion}\label{subsec:exp-results}

\begin{table*}[t]
\caption{\textbf{Empirical Results on RealCause-Lalonde.}
Metrics are aggregated across 10 random seeds (realizations). PEHE and ATE Relative Error are reported as $\text{mean} \pm \text{standard error}$. Median total wall-clock time, including training and hyperparameter optimization for traditional estimators, is reported as CATE Runtime. All times are reported using matched CPU hardware. Avg. rank is calculated over Lalonde-CPS and Lalonde-PSID combined. Best single result per column in \textbf{bold}. $^{\dagger}$IPW predicts at population level and does not produce individual-level CATE estimates.}
\centering
\small
\setlength{\tabcolsep}{2pt}
\resizebox{\textwidth}{!}{%
\begin{tabular}{l ccc | ccc | c}
\toprule
 & \multicolumn{3}{c|}{\textbf{PEHE ($\times 10^3, \downarrow$ better)}} & \multicolumn{3}{c|}{\textbf{ATE Relative Error ($\downarrow$ better)}} & \textbf{CATE Runtime (s)} \\
\cmidrule(lr){2-4} \cmidrule(lr){5-7} \cmidrule(lr){8-8}
\textbf{Method} & \textbf{Lalonde}$_{\text{CPS}}$ & \textbf{Lalonde}$_{\text{PSID}}$ & \textbf{Avg. Rank} & \textbf{Lalonde}$_{\text{CPS}}$ & \textbf{Lalonde}$_{\text{PSID}}$ & \textbf{Avg. Rank} & \textbf{Median} \\
\midrule
\multicolumn{8}{l}{\emph{Causal Foundation Models}} \\
CausalPFN & \textbf{8.97 $\pm$ 0.06} & 14.00 $\pm$ 0.41 & 1.75 $\pm$ 0.16 & 0.17 $\pm$ 0.03 & 0.24 $\pm$ 0.04 & 2.65 $\pm$ 0.26 & \phantom{00} \textbf{18.4} \\
Do-PFN & 11.96 $\pm$ 0.09 & 20.20 $\pm$ 0.39 & 4.80 $\pm$ 0.21 & 0.88 $\pm$ 0.01 & 0.89 $\pm$ 0.01 & 6.50 $\pm$ 0.30 & \phantom{0}115.1 \\
CausalFM & 12.34 $\pm$ 0.02 & 22.27 $\pm$ 0.43 & 6.60 $\pm$ 0.22 & 0.94 $\pm$ 0.00 & 0.95 $\pm$ 0.00 & 7.80 $\pm$ 0.26 & \phantom{00}31.2 \\
\midrule
\multicolumn{8}{l}{\emph{Traditional Estimators}} \\
T-Learner & 9.04 $\pm$ 0.08  & \textbf{13.65 $\pm$ 0.47} & \textbf{1.40 $\pm$ 0.11} & 0.28 $\pm$ 0.04 & \textbf{0.04 $\pm$ 0.01} & 2.25 $\pm$ 0.27 & 1803.0 \\
Debiased ML & 9.96 $\pm$ 0.34  & 15.45 $\pm$ 1.06 & 3.20 $\pm$ 0.29 & 0.34 $\pm$ 0.08 & 0.29 $\pm$ 0.08 & 3.20 $\pm$ 0.27 & 1807.9 \\
IPW$^{\dagger}$                                  & ---              & ---              & ---             & \textbf{0.15 $\pm$ 0.03} & 0.08 $\pm$ 0.02 & \textbf{2.05 $\pm$ 0.20} & --- \\
X-Learner & 11.90 $\pm$ 0.40 & 20.30 $\pm$ 0.60 & 5.20 $\pm$ 0.34 & 0.89 $\pm$ 0.06 & 0.89 $\pm$ 0.05 & 7.30 $\pm$ 0.32 & 2707.3 \\
S-Learner  & 12.45 $\pm$ 0.11 & 20.61 $\pm$ 0.41 & 5.80 $\pm$ 0.24 & 0.97 $\pm$ 0.01 & 0.71 $\pm$ 0.05 & 7.10 $\pm$ 0.34 & \phantom{0}901.4 \\
DR-Learner & 13.45 $\pm$ 0.29 & 24.09 $\pm$ 1.90 & 7.25 $\pm$ 0.28 & 0.89 $\pm$ 0.05 & 0.64 $\pm$ 0.05 & 6.15 $\pm$ 0.32 & 1820.5 \\
\bottomrule
\end{tabular}%
}
\label{tab:realcause-main}
\end{table*}

\begin{figure}[t]
\centering
\vspace{-2mm}
\includegraphics[width=0.60\linewidth]{"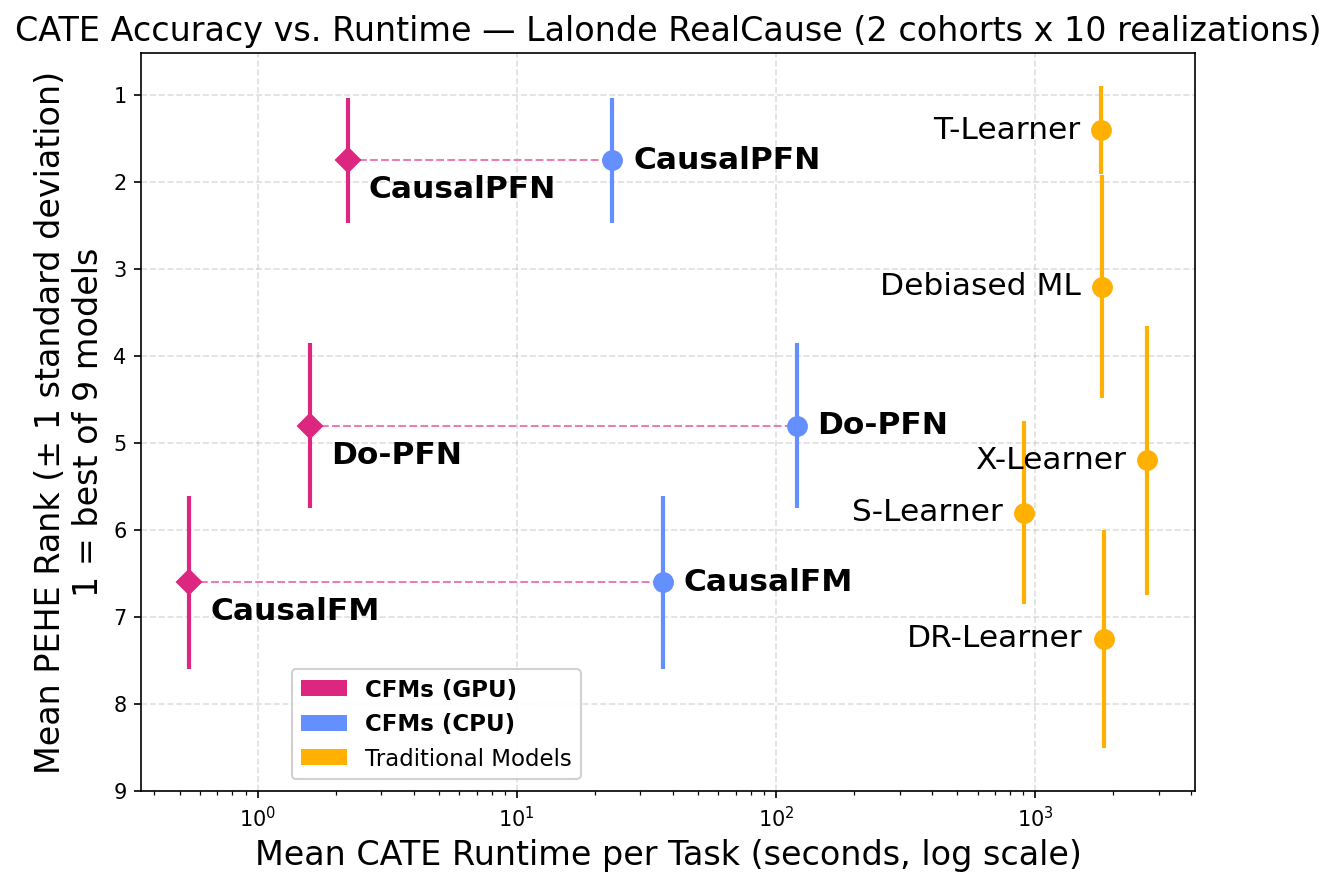"}
\vspace{-2mm}
\caption{CATE Estimation Ranking vs. Wall-Clock Runtime on RealCause-Lalonde. Causal foundation models deliver competitive estimation results while achieving orders-of-magnitude faster inference compared to tuned classical estimators because they do not need to be trained on RealCause. For CFMs, runtimes are shown both on CPU (matching the hardware used for all other models) and an A100 GPU.}
\label{fig:pehe-vs-runtime}
\vspace{-3mm}
\end{figure}

\paragraph{Benchmark Performance.} The main benchmark results are shown in \Cref{tab:realcause-main}. Despite not being trained on the Lalonde data distributions, CFMs are remarkably competitive with classical estimators. CausalPFN achieves the lowest average rank among CFMs, closely trailing the extensively tuned T-Learner baseline while outperforming it on Lalonde-CPS. IPW, a method specialized for ATE prediction that does not give individual-level predictions, showed the strongest results on ATE Relative Error $\lambda$. We note that our CausalPFN results independently replicate the values reported by \citet{balazadeh2026causalpfn}, although our T-learner and IPW baselines appear slightly stronger than in that work. CFMs that output the CID-PPD or CDTE-PPD rather than the CEPO-PPD were somewhat less performant, but still competitive with tuned X- and S-Learners. See \Cref{fig:pehe-vs-runtime} for a comparative plot of model performance.

\paragraph{Inference Amortization vs. Tuning Overhead.} CFMs show a significant runtime advantage since inference can be performed directly on raw data without training. On CPU, all three CFMs tested were 1-2 orders of magnitude faster than training, tuning, and inference with a traditional model, with CausalPFN running fastest on CPU. Further speedups in CFM runtime are obtained on GPU. CFMs have amortized the effort needed for Bayesian inference via pretraining, which we exclude from the runtime measurement since a practitioner does not need to perform it. Amortization thus represents a speedup of up to $100\times$ to get predictions of comparable quality to a highly optimized T-Learner. In production workflows with repeated evaluation across subsets, CFMs effectively eliminate the computational barrier of hyperparameter tuning and cross-fitting loops while preserving competitive CATE accuracy; see \Cref{fig:pehe-vs-runtime}.

\paragraph{Average Treatment Effect Magnitude Recovery.}

All three CFMs receive identically standardized covariates and outcomes. Under this common treatment, CausalPFN recovers the population effect closely (ATE relative error $0.17$ on Lalonde-CPS), while Do-PFN and CausalFM do not: both remain near a relative error of 1 on both cohorts ($0.88$ and $0.94$), recovering only a small fraction of the true contrast. Their errors are moreover almost identical across the two cohorts and carry very small standard errors, which points to a systematic shrinkage of the predicted effect toward zero rather than to unstable estimation.

\section{Broader Directions and Applications}\label{sec:cfms-broad}
The first generation of CFMs \citep{robertson2025do-pfn, balazadeh2026causalpfn, ma2026causalfm} established that Bayesian causal inference could be amortized across tasks, and thus that foundation models for causal inference were possible and, indeed, yielded top performance on common benchmarks. These models focus primarily on causal effect estimation in the binary treatment setting, and accept only tabular data as input. Subsequent work has broadened this initial formulation along several axes. Newer models address partial identification of causal effects, target additional treatment and data regimes, and extend beyond causal effect estimation to causal discovery and domain-specific interventional questions.

Our working definition in \Cref{def:cfm} reserves the term \emph{causal foundation model} for PFNs that estimate causal quantities, such as the CEPO, CID, or CDTE. In this section, we first review important precursor work before discussing further progress in this core causal inference setting. We then adopt a broader lens, considering closely related models for causal discovery as well as related scientific tasks. Not every model discussed below is therefore a CFM under \Cref{def:cfm}; the works we discuss in this section, however, show how the core ideas of CFMs are being applied in neighbouring areas.

\subsection{Previous work}

Prior to the first generation of CFMs, important precursor work applied the idea of learning-to-learn to causal inference, including both CaML \citep{nilforoshan2024zeroshotcausallearning} and BBCI \citep{bynum2025bbci}. These works proposed algorithms (rather than specific models) which could in principle be applied to any trainable family of models. In particular, BBCI is an algorithm which trains models to predict causal estimands on synthetic DGPs modeled by SCMs and can be deployed on unseen data. However, the class of SCMs that \citet{bynum2025bbci} train on are comparatively narrow and low-dimensional. Due to this restricted setting, there is no indication that this method will scale, which limits use for real-world, big-data environments. Causal Inference with Attention (CInA) \citep{zhang2024causalfoundationmodelduality} is another ideological precursor to modern CFMs which demonstrated a theoretical link between attention and causal inference and proposed a zero-shot, transformer-based model for causal inference.

\subsection{Extending causal inference capabilities}

\subsubsection{Incorporating structural knowledge}
The first generation of CFMs accept raw observational tabular data as input, and are unable to condition on additional domain knowledge such as partial knowledge of the underlying causal structure. For example, a practitioner may know that a covariate $X_1$ is a direct cause of $X_2$, yet the model may still assign positive posterior mass to DGPs in which $X_2$ is a cause of $X_1$. \citet{reuter2026use} address this limitation, arguing that the proper method for conditioning CFMs on additional knowledge is to utilize partial ancestral information, meaning (potentially incomplete) knowledge of the set of ancestors of each variable. They investigate different ways for CFMs to leverage this knowledge, finding empirically that learnable attention biases are particularly effective. Experiments show that conditioning on partial ancestral knowledge leads to sizeable performance gains \citep{reuter2026use}.

TabPFN-CFM \citep{zhu2026causalfoundationmodelforstructure} predicts both causal structure as well as effects. If the underlying causal graph is known, it can be added to the model inputs to improve its causal effect estimates. TabPFN-CFM is designed to predict observational, interventional, and counterfactual outcomes, thus operating at each level of Pearl's causal hierarchy \citep{pearl2009causality}.

\subsubsection{New treatment and temporal regimes}
Other work extends CFMs beyond binary treatment and time-independent data. \citet{stith2026causalfoundationmodelscontinuous} introduce CCPFN, the first CFM for continuous treatments, which reconstructs an entire individual treatment-response curve over the treatment range. This requires constructing a higher-dimensional prior over treatment and outcome mechanisms than in the binary or multi-arm setting.

Along the other axis, CausalLongPFN \citep{zare2026causallongitudinalpriorfittednetworks} is a CFM designed for potential outcome prediction in longitudinal settings with time-dependent confounding, nonlinear evolution, and other dynamical effects. At inference time, given historical data  trajectories and a proposed future treatment sequence, it estimates future counterfactual outcomes. 

There is a related line of work developing priors for temporal causal inference. CausalTimePrior \citep{thumm2026interventionaltimeseries, thumm2026causalfoundationmodelsfortimeseries} construct methods for sampling from time-dependent SCMs coupled with observational and interventional time series. \citet{thumm2026continuoustimecausalfoundationmodels} propose and construct a continuous-time version based on stochastic processes.

\subsubsection{Partial identification and sensitivity analysis} More recent work focuses on partially identifiable settings, where the treatment effect cannot be computed exactly, but rather can only be known to lie in some finite interval. IV-ICL \citep{balazadeh2026iviclboundingcausaleffects} addresses partial identifiability in the IV setting (see \Cref{subsec:data_generating_processes_and_identifiability}) where hidden confounders may exist. It estimates the PPD of the causal effect and returns bounds on the effect from the quantiles of this PPD. It provides more reliable bounds compared to traditional baselines while substantially reducing inference time. \citet{bellot2026foundationmodelspartialcausal} also build a PFN that targets partially identifiable settings, without focusing on IV specifically.

Related to this, causal sensitivity analysis studies how derived treatment effects would change in the presence of hidden confounders of different strengths \citep{rosenbaum2002observational}. \citet{javurek2026amortizingcausalsensitivityanalysis} create a PFN for causal sensitivity analysis which outputs sensitivity bounds directly. Together, these works extend CFMs to regimes where the treatment effect is not precisely identifiable, but where quantitative statements can still be made.

\subsubsection{Reliability and calibration}

A complementary line of work has developed which studies the quality and potential bias of CFM predictions. These studies reinforce the importance of a highly diverse prior for CFM pretraining. 

\citet{melnychuk2026frequentistconsistencypriordatafitted} investigate the frequentist consistency of CFM-based ATE estimation, finding that some existing CFMs have a prior-induced bias, which they heuristically note is the result of the CFM's prior being supported on DGPs with low confounding. They also introduce a one-step posterior correction method to eliminate this bias. \citet{mourao2026priordatafittednetworkscausal} show signs of low coverage of CausalPFN's credible intervals, but it is not clear whether they performed these experiments after calibrating CausalPFN \citep{balazadeh2026causalpfn}. \citet{wang2026unveilingpriordatafittednetworks} propose a task-specific fine-tuning strategy to correct bias due to low prior coverage in CFM pretraining. \citet{ham2026astructuralviewofquery} study how CFMs err when a post-treatment covariate is passed to the model at inference time, and propose a method for filtering out such inputs.

\subsection{Foundation models for causal discovery}

Causal discovery is the practice of discovering causal relationships between variables from observational data \citep{glymour2019reviewofcausaldiscoverymethods, zanga2022survey}. This runs complementary to causal inference, which focuses on the effect that interventions on one variable (treatment) have on another (outcome). Causal discovery can be formulated as a graph discovery problem where one attempts to recover the underlying causal graph relating the observed variables (\Cref{subsec:scms}). As in causal inference, a central difficulty in the field is identifiability (\Cref{subsec:data_generating_processes_and_identifiability}). In general, observational data can only identify the Markov equivalence class of the underlying DAG rather than a unique graph \citep{kalish2007estimatinghighdimensional, neal2020introduction}.

Causal discovery is a natural target for prior-fitted networks. A prior over SCMs induces a prior over causal graphs, allowing each synthetic training dataset to be paired with the graph of the SCM that generated it. A PFN can therefore be trained to map a dataset directly to a posterior distribution over causal structures, amortizing the causal discovery process. Indeed, it is argued that CFMs for causal effect estimation do so implicitly \citep{balazadeh2026causalpfn}. Several models have been created recently which could be called causal discovery foundation models.

\subsubsection{Early approaches} CSIvA \citep{ke2022learninginducecausalstructure} and AVICI \citep{lorch2022amortizedinferencecausalstructure} were early applications of neural networks and transformers to amortized causal discovery. Both models learn to directly map data (observational or interventional) to a causal structure, although only AVICI learns a posterior over causal graphs. \citet{montagna2026demystifyingamortizedcausaldiscovery} show that CSIvA generalizes poorly to unseen causal structures and demonstrates that training models on a wide variety of causal structures can improve generalization, which is aligned with the core insight of synthetic prior design in PFN and CFM training.

The Bayesian Causal Neural Process of \citet{anish2025metalearningapproachtobayesiancausaldiscovery} amortizes Bayesian causal discovery, learning to approximate a posterior over causal structures, and can generate samples from its learned posterior at inference time. However, it differs from our definition of CFMs by training on multiple smaller priors with different parameters, rather than a single highly-diverse training prior. Their experiments do show that training their model on a combination of their smaller, separate priors retains the original model's performance on each dataset. This foreshadows the development of a more general causal discovery foundation model trained on a single large prior. 

\citet{sypniewski2025amortizedcausaldiscoverypriorfitted} inserts TabPFN into the causal discovery pipeline, using it to amortize likelihood estimation for causal graphs, rather than training a new model for causal discovery. SEA \citep{wu2025sampleestimateaggregaterecipe} is an early but distinct approach to causal discovery foundation models which also trains on large amounts of synthetic data. However, it does not learn a mapping from datasets to a posterior distribution over causal structures. Rather, at inference time, SEA trains an aggregator model that takes the results from weak causal discovery predictors and summary statistics and outputs a final predicted causal graph. ADAG \citep{yin2025learningcausalgraphsscale} is an early approach which uses a transformer to directly learn a mapping from data to causal graph. However, ADAG is only seen to generalize well to unseen tasks whose underlying graphical structure or topological ordering is shared with its training data.

\subsubsection{General purpose causal discovery foundation models}

Several more recent models step more clearly into the foundation model paradigm. Arrow \citep{thompson2026arrowfoundationmodelcausal} is a zero-shot causal discovery model that operates on observational tabular data and was trained on a large and highly diverse synthetic prior over causal graphs. However, Arrow is not a Bayesian model, and outputs a single predicted DAG rather than a posterior distribution. TabCausal \citep{li2026tabcausalpretrainingcausalenvironments} is a causal discovery foundation model which operates on observational and interventional tabular data and was also trained on a large, diverse synthetic prior. TabCausal is also not a Bayesian learner, outputting only a probabilistic adjacency matrix representing the predicted directed edge likelihoods given the input data. Similarly, FoundCause \citep{blobaum2026foundcausecausaldiscoverylatent} is a foundation model which outputs a probabilistic adjacency matrix, as well as a confounding probability matrix representing the probability that any two variables are confounded by a hidden variable.

DCD-PFN \citep{guan2026dcdpfndecouplingawarefoundationmodel} applies a local-to-global method, first learning to approximate the posterior distribution of the Markov boundary of target variables and later patching these together to form a global causal graph. TabPFN-CFM \citep[mentioned above]{zhu2026causalfoundationmodelforstructure} is trained to perform both causal discovery and causal inference. When the causal graph is unknown, the model learns an approximation to the posterior distribution of the true causal graph. CDFM \citep{qiao2026cdfmgeneralpurposecausaldiscovery} is another causal discovery foundation model trained on a large, diverse synthetic prior, which generalizes to unseen evaluation scenarios. DAG-FM \citep{chen2026dagfmfoundationmodelcausal} is a foundation model using two transformer sub-modules, one for leaf nodes and one for parent nodes, combining the outputs of both to output a prediction of the true causal graph. DAG-FM is also pretrained on a large, diverse synthetic prior.

Temporal causal discovery has also been addressed through large-scale synthetic pretraining. Though not a Bayesian learner, \citet{kougioulis2026largecausalmodelstemporal} created a model for temporal causal discovery which directly predicts a time-dependent adjacency matrix rather than a posterior distribution. They train on a prior which includes time-dependent SCMs constructed from real time series data, in addition to purely synthetic graph-sampling methods.

\subsection{Domain-specific and adjacent applications}

The same prior-fitting methodology employed by CFMs can be applied to more focused scientific domains where causal effect estimation for specific yet highly complex mechanisms is studied. Due to the high complexity of these domains, a synthetic prior constructed specially for each can be diverse enough to justify pretraining at scale.

\subsubsection{Single-cell perturbation models}

Single-cell perturbation modeling is a central topic in biology which aims to estimate the molecular response to perturbations within a single cell \citep{bunne2023learningsinglecell}. MapPFN \citep{sextro2026mappfnlearningcausalperturbation} is a CFM trained on a synthetic biological prior which predicts the post-perturbation distribution of single-cell systems. Their model takes a set of observational data and interventional experiments, and outputs an approximation to the post-perturbation distribution of a new intervention. PerturbPFN \citep{gao2026perturbpfnprobinglimitssynthetic} is another perturbation effect estimator which operates differently. It first estimates an underlying causal graph which, coupled with an SCM decoder, outputs an estimate of the post-perturbation distribution. Both models demonstrate strong performance compared to baselines in the field.

\subsubsection{Survival analysis}

Survival analysis deals with predicting the time-to-event for different events of interest \citep{klein2003survival}. This could be the time between diagnosis and patient death \citep{clark2003survivalanalysispartibasicconcepts}, the time until a user churns \citep{vandenpoel2004customerattritionanalysis}, or the time until a device fails \citep{meeker1998statisticalmethods}. A central issue in this field is that data are often censored or truncated, meaning that data is only partially known or excluded from the observed sample. Recent works have introduced PFNs to survival analysis, creating foundational models like SIC \citep{seletkov2026survivalincontextamortizedbayesian}, SurvivalPFN \citep{qi2026survivalpfnamortizingsurvivalprediction}, and SurvPFN \citep{bohm2026survpfnfoundationmodelssurvival}. These all pretrain a PFN on a prior over synthetic survival analysis DGPs. SurvivalPFN and SIC both achieve top performance compared to a wide variety of non-foundational survival models, including classical survival models, tree-based models, and neural models, and SurvPFN is competitive across all evaluated benchmarks. Interestingly, SurvivalPFN also outperforms work by \citet{kim2026tabularfoundationmodelssurvival}, an approach which casts survival analysis as a binary classification problem that it passes to a TFM. Although this enables off-the-shelf TFMs to be applied to survival analysis, this demonstrates a clear advantage in SurvivalPFN's pretraining process. 

\subsection{Future directions}

The models discussed above point to encouraging trends in CFM development and broader usage. We have seen a wide variety of foundation models applied across various domains. These suggest several areas for future improvements.

Existing CFMs are typically specialized along three distinct axes: the treatment regime, the identification assumptions, and the target causal quantity. A natural future goal is a single CFM that supports multiple treatment regimes, different identification assumptions, and estimation of many causal quantities at once. Such a model would support binary, multi-arm, and continuous treatment (and combinations thereof); be deployed in backdoor, IV, or other settings; and be used to estimate the CEPO, CID, and CDTE, in addition to other estimands. Increased generality should not come at the cost of hiding identification assumptions; a useful interface would allow the analyst to specify the estimand as well as the assumptions, indicate whether the estimand is identified under those assumptions, and propagate uncertainty about causal structure into downstream estimation and uncertainty quantification. One or more common internal representations could then support several downstream tasks without requiring a separate pretrained model for each. Recent models which perform both causal discovery and inference provide an early step in this direction. 

Instead of operating on a single observational tabular dataset, future CFMs could incorporate more diverse information sources, such as experimental datasets, partial causal knowledge, or domain constraints. Recent models which incorporate partial ancestral knowledge or graphical structure are promising examples. More flexible interfaces might also allow a fixed model to revise its estimates through additional context if new experiments or structural knowledge become available.

The quality of the synthetic training prior directly affects CFM performance and generalization. Developing more quantitative measures of causal prior coverage would help improve prior design. Along this direction, useful properties of future CFMs would include automatic detection of mismatch between the target DGP and the training prior. Both problems are especially difficult in the causal setting, where observational data does not fully characterize the distributions of interest.
\section{Conclusion}\label{sec:conclusion}
Causal foundation models are an emerging approach to solving problems in causal inference, and are beginning to be applied to different scientific domains. They are transformer-based models that are pretrained once on a synthetic causal prior; at inference time, they leverage in-context learning to estimate causal effects in unseen settings. This paradigm drastically cuts down on model deployment time compared to traditional estimators which have to be trained and tuned for each new problem. At the same time, CFMs demonstrate competitive performance on causal benchmarks compared to traditional estimators.

In this work, we provide a hands-on introduction to CFMs. We include several Jupyter notebooks  (\jupyter{https://github.com/layer6ai-labs/cfms/tree/main/notebooks}) in order to help the reader use these models immediately, as well as empirical results showing the effectiveness and speed of current CFMs. Our hope is that this introduces CFMs to a wider audience and encourages the reader to get started using CFMs in their own work.

\subsubsection*{Broader Impact Statement}
The nature of our work is an introduction to emerging techniques in causal foundation modeling. Our aim is to demystify how these methods operate and bring new techniques to a wider audience. As such, we do not anticipate broader societal repercussions from this work.

\bibliographystyle{tmlr}
\bibliography{main.bib}

@article{bynum2025bbci,
      title={{Black Box Causal Inference: Effect Estimation via Meta Prediction}}, 
      author={Lucius E. J. Bynum and Aahlad Manas Puli and Diego Herrero-Quevedo and Nhi Nguyen and Carlos Fernandez-Granda and Kyunghyun Cho and Rajesh Ranganath},
  journal={arXiv:2503.05985},
  year={2025}
}

@inproceedings{robertson2025do-pfn,
 author = {Robertson, Jake and Reuter, Arik and Guo, Siyuan and Hollmann, Noah and Hutter, Frank and Sch\"{o}lkopf, Bernhard},
 booktitle = {Advances in Neural Information Processing Systems},
 pages = {174811--174848},
 title = {{Do-PFN: In-Context Learning for Causal Effect Estimation}},
 volume = {38},
 year = {2025}
}

@inproceedings{balazadeh2026causalpfn,
  title={{CausalPFN: Amortized Causal Effect Estimation via In-Context Learning}},
  booktitle={Advances in Neural Information Processing Systems},
  author={Balazadeh, Vahid and Kamkari, Hamidreza and Thomas, Valentin and Ma, Junwei and Li, Bingru and Cresswell, Jesse C. and Krishnan, Rahul G.},
  volume={38},
  pages={154945--154984},
  year={2025}
}

@inproceedings{ma2026causalfm,
 author = {Ma, Yuchen and Frauen, Dennis and Javurek, Emil and Feuerriegel, Stefan},
 booktitle = {International Conference on Learning Representations},
 pages = {79065--79098},
 title = {Foundation Models for Causal Inference via Prior-Data Fitted Networks},
 volume = {2026},
 year = {2026}
}

@misc{neal2020introduction,
  title        = {Introduction to Causal Inference from a Machine Learning Perspective},
  author       = {Neal, Brady},
  year         = {2020},
  howpublished = {\url{https://www.bradyneal.com/Introduction_to_Causal_Inference-Dec17_2020-Neal.pdf}},
}

@article{chernozhukov2018double,
author = {Chernozhukov, Victor and Chetverikov, Denis and Demirer, Mert and Duflo, Esther and Hansen, Christian and Newey, Whitney and Robins, James},
title = {Double/debiased machine learning for treatment and structural parameters},
journal = {The Econometrics Journal},
volume = {21},
number = {1},
doi = {10.1111/ectj.12097},
year = {2018}
}

@article{wager2018,
author = {Stefan Wager and Susan Athey},
title = {Estimation and Inference of Heterogeneous Treatment Effects using Random Forests},
journal = {Journal of the American Statistical Association},
volume = {113},
number = {523},
pages = {1228--1242},
year = {2018},
publisher = {Taylor \& Francis},
doi = {10.1080/01621459.2017.1319839},
}

@article{zhang2024causalfoundationmodelduality,
      title={{Towards Causal Foundation Model: on Duality between Causal Inference and Attention}}, 
      author={Jiaqi Zhang and Joel Jennings and Agrin Hilmkil and Nick Pawlowski and Cheng Zhang and Chao Ma},
  journal={arXiv:2310.00809},
  year={2023}
}

@inproceedings{nilforoshan2024zeroshotcausallearning,
 author = {Nilforoshan, Hamed and Moor, Michael and Roohani, Yusuf and Chen, Yining and \v{S}urina, Anja and Yasunaga, Michihiro and Oblak, Sara and Leskovec, Jure},
 booktitle = {Advances in Neural Information Processing Systems},
 pages = {6862--6901},
 title = {{Zero-shot causal learning}},
 volume = {36},
 year = {2023}
}

@inproceedings{
hollmann2023tabpfn,
title={{TabPFN: A Transformer That Solves Small Tabular Classification Problems in a Second}},
author={Noah Hollmann and Samuel M{\"u}ller and Katharina Eggensperger and Frank Hutter},
booktitle={The Eleventh International Conference on Learning Representations },
year={2023},
}

@InProceedings{qu2025tabicl,
  title = 	 {{TabICL: A Tabular Foundation Model for In-Context Learning on Large Data}},
  author =       {Qu, Jingang and Holzm\"{u}ller, David and Varoquaux, Ga\"{e}l and Le Morvan, Marine},
  booktitle = 	 {Proceedings of the 42nd International Conference on Machine Learning},
  pages = 	 {50817--50847},
  year = 	 {2025},
  volume = 	 {267},
  series = 	 {Proceedings of Machine Learning Research},
}

@inproceedings{ma2025tabdpt,
      title={{TabDPT: Scaling Tabular Foundation Models on Real Data}}, 
      author={Junwei Ma and Valentin Thomas and Rasa Hosseinzadeh and Alex Labach and Hamidreza Kamkari and Jesse C. Cresswell and Keyvan Golestan and Guangwei Yu and Anthony L. Caterini and Maksims Volkovs},
  booktitle={Advances in Neural Information Processing Systems},
  year={2025}
}

@book{imbens2015causalinference, place={Cambridge}, title={Causal Inference for Statistics, Social, and Biomedical Sciences: An Introduction}, publisher={Cambridge University Press}, author={Imbens, Guido W. and Rubin, Donald B.}, year={2015}}

@article{javurek2026amortizingcausalsensitivityanalysis,
      title={Amortizing Causal Sensitivity Analysis via Prior Data-Fitted Networks}, 
      author={Emil Javurek and Dennis Frauen and Marie Brockschmidt and Jonas Schweisthal and Stefan Feuerriegel},
  journal={arXiv:2605.10590},
  year={2026}
}

@book{pearl2009causality, 
place={Cambridge}, 
edition={2nd},
title={Causality},
publisher={Cambridge University Press},
author={Pearl, Judea},
year={2009}}

@article{mooij2016distinguishingcausefromeffect,
  author  = {Joris M. Mooij and Jonas Peters and Dominik Janzing and Jakob Zscheischler and Bernhard Sch{{\"o}}lkopf},
  title   = {{Distinguishing Cause from Effect Using Observational Data: Methods and Benchmarks}},
  journal = {Journal of Machine Learning Research},
  year    = {2016},
  volume  = {17},
  number  = {32},
  pages   = {1--102},
}

@article{matthews2000storks,
  author  = {Matthews, Robert},
  title   = {Storks Deliver Babies ($p = 0.008$)},
  journal = {Teaching Statistics},
  volume  = {22},
  pages   = {36--38},
  year    = {2000},
  doi     = {10.1111/1467-9639.00013}
}

@inproceedings{vaswani2017attention,
 author = {Vaswani, Ashish and Shazeer, Noam and Parmar, Niki and Uszkoreit, Jakob and Jones, Llion and Gomez, Aidan N and Kaiser, \L ukasz and Polosukhin, Illia},
 booktitle = {Advances in Neural Information Processing Systems},
 title = {Attention is All you Need},
 volume = {30},
 year = {2017}
}

@article{rubin2005causalinference,
author = {Donald B Rubin},
title = {{Causal Inference Using Potential Outcomes}},
journal = {Journal of the American Statistical Association},
volume = {100},
number = {469},
pages = {322--331},
year = {2005},
publisher = {Taylor \& Francis},
doi = {10.1198/016214504000001880},
}

@article{pear1995causaldiagrams,
 ISSN = {00063444, 14643510},
 author = {Judea Pearl},
 journal = {Biometrika},
 number = {4},
 pages = {669--688},
 publisher = {[Oxford University Press, Biometrika Trust]},
 title = {{Causal Diagrams for Empirical Research}},
 volume = {82},
 year = {1995}
}

@inproceedings{huang2012pearlscalculusinterventioncomplete,
author = {Huang, Yimin and Valtorta, Marco},
title = {{Pearl's calculus of intervention is complete}},
year = {2006},
isbn = {0974903922},
publisher = {AUAI Press},
booktitle = {Proceedings of the Twenty-Second Conference on Uncertainty in Artificial Intelligence},
pages = {217–224},
numpages = {8}
}

@inproceedings{shpitser2006identification,
author = {Shpitser, Ilya and Pearl, Judea},
title = {Identification of joint interventional distributions in recursive semi-{M}arkovian causal models},
year = {2006},
isbn = {9781577352815},
booktitle = {Proceedings of the 21st National Conference on Artificial Intelligence - Volume 2},
pages = {1219–1226},
numpages = {8},
location = {Boston, Massachusetts},
series = {AAAI'06}
}

@article{yao2021surveyoncausalinference,
author = {Yao, Liuyi and Chu, Zhixuan and Li, Sheng and Li, Yaliang and Gao, Jing and Zhang, Aidong},
title = {{A Survey on Causal Inference}},
year = {2021},
publisher = {Association for Computing Machinery},
volume = {15},
number = {5},
issn = {1556-4681},
doi = {10.1145/3444944},
journal = {ACM Trans. Knowl. Discov. Data},
month = may,
articleno = {74},
numpages = {46},
}

@inproceedings{
muller2022transformers,
title={{Transformers Can Do Bayesian Inference}},
author={Samuel M{\"u}ller and Noah Hollmann and Sebastian Pineda Arango and Josif Grabocka and Frank Hutter},
booktitle={International Conference on Learning Representations},
year={2022},
}

@article{tabpfn_v3,
      title={Tab{PFN}-3: Technical Report}, 
      author={Léo Grinsztajn and Klemens Flöge and Oscar Key and Felix Birkel and Philipp Jund and Brendan Roof and Mihir Manium and Shi Bin Hoo and Magnus Bühler and Anurag Garg and Dominik Safaric and Jake Robertson and Benjamin Jäger and Simone Alessi and Adrian Hayler and Vladyslav Moroshan and Lennart Purucker and Philipp Singer and Alan Arazi and Julien Siems and Jan Hendrik Metzen and Georg Grab and Nick Erickson and Siyuan Guo and Eliott Kalfon and Simon Bing and David Salinas and Clara Cornu and Lilly Charlotte Wehrhahn and Diana Kriuchkova and Kursat Kaya and Lydia Sidhoum and Marie Salmon and Jerry Chen and Madelon Hulsebos and Yann LeCun and Samuel Müller and Bernhard Schölkopf and Sauraj Gambhir and Noah Hollmann and Frank Hutter},
  journal={arXiv:2605.13986},
  year={2026}
}

@InProceedings{qu2026tabiclv2betterfasterscalable,
      title={{TabICLv2}: A better, faster, scalable, and open tabular foundation model}, 
      author={Jingang Qu and David Holzmüller and Gaël Varoquaux and Marine Le Morvan},
  booktitle = {Proceedings of the 43rd International Conference on Machine Learning},
  year = 	 {2026},
  series = 	 {Proceedings of Machine Learning Research},
}

@article{hosseinzadeh2026tabdptturbo,
  title={{TabDPT-Turbo: Efficient In-Context Learning for Tabular Prediction}},
  author={Hosseinzadeh, Rasa and Labach, Alex and Xue, Zexin and Han, Shuyi and Thomas, Valentin and Caterini, Anthony L},
  journal={arXiv:2608.01400},
  year={2026}
}

@article{holland1986statisticsandcausalinference,
 ISSN = {01621459, 1537274X},
 author = {Paul W. Holland},
 journal = {Journal of the American Statistical Association},
 number = {396},
 pages = {945--960},
 publisher = {[American Statistical Association, Taylor & Francis, Ltd.]},
 title = {{Statistics and Causal Inference}},
 volume = {81},
 year = {1986}
}

@article{balazadeh2026iviclboundingcausaleffects,
      title={{IV-ICL: Bounding Causal Effects with Instrumental Variables via In-Context Learning}}, 
      author={Vahid Balazadeh and Hamidreza Kamkari and Medha Barath and Ricardo Silva and Rahul G. Krishnan},
  journal={arXiv:2605.12924},
  year={2026} 
}

@book{manski2003partial,
  author    = {Manski, Charles F.},
  title     = {Partial Identification of Probability Distributions},
  series    = {Springer Series in Statistics},
  publisher = {Springer},
  address   = {New York, NY},
  year      = {2003},
  isbn      = {978-0-387-00454-9}
}

@article{schwab2020doseresponse,
  title     = {{Learning Counterfactual Representations for Estimating Individual Dose-Response Curves}},
  author    = {Schwab, Patrick and Linhardt, Lorenz and Bauer, Stefan and Buhmann, Joachim M. and Karlen, Walter},
  journal   = {Proceedings of the AAAI Conference on Artificial Intelligence},
  volume    = {34},
  number    = {04},
  pages     = {5612--5619},
  year      = {2020},
  doi       = {10.1609/aaai.v34i04.6014},
}

@article{imbens2020potentialoutcomeanddirectedacyclicgraph,
Author = {Imbens, Guido W.},
Title = {{Potential Outcome and Directed Acyclic Graph Approaches to Causality: Relevance for Empirical Practice in Economics}},
Journal = {Journal of Economic Literature},
Volume = {58},
Number = {4},
Year = {2020},
Pages = {1129–79},
DOI = {10.1257/jel.20191597},
}

@article{pearl2009causalinferenceinstatisticsanoverview,
author = {Judea Pearl},
title = {{Causal inference in statistics: An overview}},
volume = {3},
journal = {Statistics Surveys},
publisher = {Amer. Statist. Assoc., the Bernoulli Soc., the Inst. Math. Statist., and the Statist. Soc. Canada},
pages = {96 -- 146},
year = {2009},
doi = {10.1214/09-SS057},
}

@article{chipman2010bart,
   title={{BART: Bayesian additive regression trees}},
   volume={4},
   ISSN={1932-6157},
   DOI={10.1214/09-aoas285},
   number={1},
   journal={The Annals of Applied Statistics},
   publisher={Institute of Mathematical Statistics},
   author={Chipman, Hugh A. and George, Edward I. and McCulloch, Robert E.},
   year={2010},
}

@inproceedings{alaa2017bayesianinferenceofindividualized,
 author = {Alaa, Ahmed M. and van der Schaar, Mihaela},
 booktitle = {Advances in Neural Information Processing Systems},
 title = {Bayesian Inference of Individualized Treatment Effects using Multi-task {G}aussian Processes},
 volume = {30},
 year = {2017}
}

@inproceedings{brown2020icl,
 author = {Brown, Tom and Mann, Benjamin and Ryder, Nick and Subbiah, Melanie and Kaplan, Jared D and Dhariwal, Prafulla and Neelakantan, Arvind and Shyam, Pranav and Sastry, Girish and Askell, Amanda and Agarwal, Sandhini and Herbert-Voss, Ariel and Krueger, Gretchen and Henighan, Tom and Child, Rewon and Ramesh, Aditya and Ziegler, Daniel and Wu, Jeffrey and Winter, Clemens and Hesse, Chris and Chen, Mark and Sigler, Eric and Litwin, Mateusz and Gray, Scott and Chess, Benjamin and Clark, Jack and Berner, Christopher and McCandlish, Sam and Radford, Alec and Sutskever, Ilya and Amodei, Dario},
 booktitle = {Advances in Neural Information Processing Systems},
 title = {Language Models are Few-Shot Learners},
 volume = {33},
 year = {2020}
}

@article{hoffman2014nouturn,
  author  = {Matthew D. Hoffman and Andrew Gelman},
  title   = {{The No-U-Turn Sampler: Adaptively Setting Path Lengths in Hamiltonian Monte Carlo}},
  journal = {Journal of Machine Learning Research},
  year    = {2014},
  volume  = {15},
  number  = {47},
  pages   = {1593--1623},
}

@InProceedings{shalit2017estimatingindividualtreatmenteffects,
  title = 	 {Estimating individual treatment effect: {G}eneralization bounds and algorithms},
  author =       {Uri Shalit and Fredrik D. Johansson and David Sontag},
  booktitle = 	 {Proceedings of the 34th International Conference on Machine Learning},
  pages = 	 {3076--3085},
  year = 	 {2017},
  volume = 	 {70},
  series = 	 {Proceedings of Machine Learning Research},
}

@article{athey2017thestateofappliedeconometrics,
Author = {Athey, Susan and Imbens, Guido W.},
Title = {The State of Applied Econometrics: Causality and Policy Evaluation},
Journal = {Journal of Economic Perspectives},
Volume = {31},
Number = {2},
Year = {2017},
Month = {May},
Pages = {3–32},
DOI = {10.1257/jep.31.2.3}
}

@article{bottou2013counterfactualreasoningandlearning,
  author  = {L{{\'e}}on Bottou and Jonas Peters and Joaquin Qui{{\~n}}onero-Candela and Denis X. Charles and D. Max Chickering and Elon Portugaly and Dipankar Ray and Patrice Simard and Ed Snelson},
  title   = {Counterfactual Reasoning and Learning Systems: The Example of Computational Advertising},
  journal = {Journal of Machine Learning Research},
  year    = {2013},
  volume  = {14},
  number  = {101},
  pages   = {3207--3260},
}

@article{gordon2019comparisonofapproaches,
  author  = {Gordon, Brett R. and Zettelmeyer, Florian and Bhargava, Neha and Chapsky, Dan},
  title   = {A Comparison of Approaches to Advertising Measurement: Evidence from Big Field Experiments at {Facebook}},
  journal = {Marketing Science},
  year    = {2019},
  volume  = {38},
  number  = {2},
  pages   = {193--225},
  doi     = {10.1287/mksc.2018.1135},
}

@InProceedings{finn2017modelagnosticmetalearning,
  title = 	 {{Model-Agnostic Meta-Learning for Fast Adaptation of Deep Networks}},
  author =       {Chelsea Finn and Pieter Abbeel and Sergey Levine},
  booktitle = 	 {Proceedings of the 34th International Conference on Machine Learning},
  pages = 	 {1126--1135},
  year = 	 {2017},
  volume = 	 {70},
  series = 	 {Proceedings of Machine Learning Research},
}

@inbook{frauen2026machinelearningforcausalinference,
author = {Frauen, Dennis and Melnychuk, Valentyn and van der Laan, Lars and Feuerriegel, Stefan},
publisher = {John Wiley \& Sons, Ltd},
isbn = {9781118445112},
title = {Machine Learning for Causal Inference},
booktitle = {Wiley StatsRef: Statistics Reference Online},
chapter = {},
pages = {1-17},
doi = {10.1002/9781118445112.stat08670},
year = {2026},
}

@article{stith2026causalfoundationmodelscontinuous,
      title={Causal Foundation Models with Continuous Treatments}, 
      author={Christopher Stith and Medha Barath and Vahid Balazadeh and Jesse C. Cresswell and Rahul G. Krishnan},
  journal={arXiv:2605.15133},
  year={2026}
}

@inproceedings{zhang2025mitramixedsyntheticpriors,
 author = {Zhang, Xiyuan and Maddix Robinson, Danielle and Yin, Junming and Erickson, Nick and Ansari, Abdul Fatir and Han, Boran and Zhang, Shuai and Akoglu, Leman and Faloutsos, Christos and Mahoney, Michael and Hu, Tony and Rangwala, Huzefa and Karypis, George and Wang, Yuyang (Bernie)},
 booktitle = {Advances in Neural Information Processing Systems},
 doi = {10.52202/085713-0535},
 pages = {15795--15840},
 title = {Mitra: Mixed Synthetic Priors for Enhancing Tabular Foundation Models},
 volume = {38},
 year = {2025}
}

@article{turkmen2026evaluatingdatapriorstabular,
      title={{Towards Evaluating Data Priors for Tabular Foundation Models}}, 
      author={Zeynep Türkmen and Kürşat Kaya and Alexander Pfefferle and Frank Hutter},
      year={2026},
      journal={arXiv:2606.29241}, 
}

@article{sextro2026mappfnlearningcausalperturbation,
      title={{MapPFN: Learning Causal Perturbation Maps in Context}}, 
      author={Marvin Sextro and Weronika Kłos and Gabriel Dernbach},
      year={2026},
      journal={arXiv:2601.21092}, 
}

@article{zanga2022survey,
  title={{A survey on causal discovery: Theory and practice}},
  author={Zanga, Alessio and Ozkirimli, Elif and Stella, Fabio},
  journal={International Journal of Approximate Reasoning},
  volume={151},
  pages={101--129},
  year={2022},
  publisher={Elsevier}
}

@article{zhu2026causalfoundationmodelforstructure,
  title={{A Causal Foundation Model for Structure and Outcome Prediction}},
  author={Zhu, Max and Mansoldo, Martino and Wang, Ching-Hao and Groha, Stefan},
  journal={arXiv:2606.26467},
  year={2026}
}

@inproceedings{anish2025metalearningapproachtobayesiancausaldiscovery,
 author = {Dhir, Anish and Ashman, Matthew and Requeima, James and van der Wilk, Mark},
 booktitle = {International Conference on Learning Representations},
 pages = {14158--14178},
 title = {{A Meta-Learning Approach to Bayesian Causal Discovery}},
 year = {2025}
}

@inproceedings{
ke2022learninginducecausalstructure,
title={{Learning to Induce Causal Structure}},
author={Nan Rosemary Ke and Silvia Chiappa and Jane X Wang and Jorg Bornschein and Anirudh Goyal and Melanie Rey and Theophane Weber and Matthew Botvinick and Michael Curtis Mozer and Danilo Jimenez Rezende},
booktitle={International Conference on Learning Representations},
year={2023},
}

@inproceedings{lorch2022amortizedinferencecausalstructure,
 author = {Lorch, Lars and Sussex, Scott and Rothfuss, Jonas and Krause, Andreas and Sch\"{o}lkopf, Bernhard},
 booktitle = {Advances in Neural Information Processing Systems},
 doi = {10.52202/068431-0952},
 pages = {13104--13118},
 title = {{Amortized Inference for Causal Structure Learning}},
 volume = {35},
 year = {2022}
}

@article{sypniewski2025amortizedcausaldiscoverypriorfitted,
  title={{Amortized Causal Discovery with Prior-Fitted Networks}},
      author={Mateusz Sypniewski and Mateusz Olko and Mateusz Gajewski and Piotr Miłoś},
  journal={arXiv:2512.11840},
  year={2025}
}

@book{neal1996bayesian,
  title={Bayesian Learning for Neural Networks},
  author={Neal, Radford M.},
  isbn={978-0-387-94724-2},
  series={Lecture Notes in Statistics},
  doi={10.1007/978-1-4612-0745-0},
  year={1996},
  publisher={Springer New York}
}

@article{wainwright2008graphical,
  title     = {{Graphical Models, Exponential Families, and Variational Inference}},
  author    = {Wainwright, Martin J. and Jordan, Michael I.},
  journal   = {Foundations and Trends in Machine Learning},
  volume    = {1},
  number    = {1--2},
  pages     = {1--305},
  year      = {2008},
  publisher = {Now Publishers Inc},
  doi       = {10.1561/2200000001}
}

@article{jordan1999introduction,
  title     = {{An Introduction to Variational Methods for Graphical Models}},
  author    = {Jordan, Michael I. and Ghahramani, Zoubin and Jaakkola, Tommi S. and Saul, Lawrence K.},
  journal   = {Machine Learning},
  volume    = {37},
  number    = {2},
  pages     = {183--233},
  year      = {1999},
  publisher = {Springer},
  doi       = {10.1023/A:1007665907178}
}

@inproceedings{papamakarios2016fastepsilonfree,
 author = {Papamakarios, George and Murray, Iain},
 booktitle = {Advances in Neural Information Processing Systems},
 title = {{Fast $\epsilon$-free Inference of Simulation Models with Bayesian Conditional Density Estimation}},
 volume = {29},
 year = {2016}
}

@book{gelman2013bayesiandataanalysis,
  title={{Bayesian Data Analysis}},
  author={Gelman, A. and Carlin, J.B. and Stern, H.S. and Dunson, D.B. and Vehtari, A. and Rubin, D.B.},
  isbn={9781439840955},
  edition={3rd},
  series={Chapman \& Hall/CRC Texts in Statistical Science},
  year={2013},
  publisher={Taylor \& Francis}
}

@article{hollmann2025accurate,
  title={{Accurate predictions on small data with a tabular foundation model}},
  author={Hollmann, Noah and M{\"u}ller, Samuel and Purucker, Lennart and Krishnakumar, Arjun and K{\"o}rfer, Max and Hoo, Shi Bin and Schirrmeister, Robin Tibor and Hutter, Frank},
  journal={Nature},
  volume={637},
  number={8045},
  pages={319--326},
  year={2025},
  publisher={Nature Publishing Group UK London}
}

@article{hayler2026advancingopenreproduciblerelational,
      title={{Advancing Open and Reproducible Relational Learning: RelArena-$\alpha$, TabPFN-Rel and RPI}}, 
      author={Adrian Hayler and Klemens Flöge and Alan Arazi and Rishabh Ranjan and Jure Leskovec and Felix Birkel and Brendan Roof and Anurag Garg and Kristina Collins and Lydia Sidhoum and Jonas Kübler and Siyuan Guo and Oscar Key and Jan Hendrik Metzen and Rylee Grace and David Salinas and Arthur Cahu and Simon Bing and Benjamin Jäger and Tuana Çelik and Mihir Manium and Vitor Monteiro and Jake Robertson and Jerry Chen and Eliott Kalfon and Tomás Pereda and Lilly Wehrhahn and Dominik Safaric and Tobias Schroeder and Georg Grab and Diana Kriuchkova and Clara Cornu and Philipp Singer and Nick Erickson and Vahid Balazadeh and Marie Salmon and Simone Alessi and Kürşat Kaya and Philipp Jund and Léo Grinsztajn and Yann LeCun and Bernhard Schölkopf and Madelon Hulsebos and Lennart Purucker and Sauraj Gambhir and Frank Hutter and Noah Hollmann},
  journal={arXiv:2608.16319},
  year={2026}
}

@inproceedings{dooley2023forecastpfn,
 author = {Dooley, Samuel and Khurana, Gurnoor Singh and Mohapatra, Chirag and Naidu, Siddartha V and White, Colin},
 booktitle = {Advances in Neural Information Processing Systems},
 doi = {10.52202/075280-0112},
 pages = {2403--2426},
 title = {{ForecastPFN: Synthetically-Trained Zero-Shot Forecasting}},
 volume = {36},
 year = {2023}
}

@inproceedings{taga2025timepfn,
  title     = {{TimePFN: Effective Multivariate Time Series Forecasting with Synthetic Data}},
  author    = {Taga, Ege Onur and Ildiz, Muhammed Emrullah and Oymak, Samet},
  booktitle = {Proceedings of the AAAI Conference on Artificial Intelligence},
  volume    = {39},
  pages     = {20761--20769},
  year      = {2025},
  doi       = {10.1609/aaai.v39i19.34288}
}

@InProceedings{wang2026relationalincontextlearningsynthetic,
title={{Relational In-Context Learning via Synthetic Pre-training with Structural Prior}},
author={Yanbo Wang and Jiaxuan You and Chuan Shi and Muhan Zhang},
  booktitle = {Proceedings of the 43rd International Conference on Machine Learning},
  year = 	 {2026},
  series = 	 {Proceedings of Machine Learning Research},
}

@article{lalonde1986evaluating,
  title={{Evaluating the Econometric Evaluations of Training Programs with Experimental Data}},
  author={LaLonde, Robert J},
  journal={The American Economic Review},
  volume={76},
  number={4},
  pages={604--620},
  year={1986},
  publisher={JSTOR}
}

@article{neal2020realcause,
  title={{RealCause: Realistic Causal Inference Benchmarking}},
  author={Neal, Brady and Huang, Chin-Wei and Raghupathi, Sunand},
  journal={arXiv:2011.15007},
  year={2020}
}

@article{
kunzel2019metalearners,
author = {Sören R. Künzel  and Jasjeet S. Sekhon  and Peter J. Bickel  and Bin Yu},
title = {{Metalearners for estimating heterogeneous treatment effects using machine learning}},
journal = {Proceedings of the National Academy of Sciences},
volume = {116},
number = {10},
pages = {4156-4165},
year = {2019},
doi = {10.1073/pnas.1804597116}
}

@article{robins1994estimation,
  title={Estimation of regression coefficients when some regressors are not always observed},
  author={Robins, James M and Rotnitzky, Andrea and Zhao, Lue Ping},
  journal={Journal of the American Statistical Association},
  volume={89},
  number={427},
  pages={846--866},
  year={1994},
  publisher={Taylor \& Francis}
}

@article{kennedy2023towards,
  title={Towards optimal doubly robust estimation of heterogeneous causal effects},
  author={Kennedy, Edward H},
  journal={Electronic Journal of Statistics},
  volume={17},
  number={2},
  pages={3008--3049},
  year={2023},
  publisher={Institute of Mathematical Statistics and Bernoulli Society}
}

@misc{battocchi2019econml,
  author={Keith Battocchi and Eleanor Dillon and Maggie Hei and Greg Lewis and Paul Oka and Miruna Oprescu and Vasilis Syrgkanis},
  title={{EconML}: {A Python Package for ML-Based Heterogeneous Treatment Effects Estimation}},
  howpublished={https://github.com/py-why/EconML},
  note={Version 0.15.0},
  year={2019}
}

@article{zare2026causallongitudinalpriorfittednetworks,
      title={{Causal Longitudinal Prior-Fitted Networks for Counterfactual Outcome Prediction}}, 
      author={Amirhossein Zare and Amirhessam Zare and Herlock Rahimi and Reza Salarikia and Mohammad Kashkooli},
  journal={arXiv:2606.05797},
  year={2026} 
}

@book{klein2003survival,
  title     = {{Survival Analysis: Techniques for Censored and Truncated Data}},
  author    = {Klein, John P. and Moeschberger, Melvin L.},
  year      = {2003},
  edition   = {2nd},
  publisher = {Springer},
  address   = {New York},
  doi       = {10.1007/b97377}
}

@article{clark2003survivalanalysispartibasicconcepts,
  title   = {{Survival Analysis Part I: Basic Concepts and First Analyses}},
  author  = {Clark, T. G. and Bradburn, M. J. and
             Love, S. B. and Altman, D. G.},
  journal = {British Journal of Cancer},
  volume  = {89},
  pages   = {232--238},
  year    = {2003},
  doi     = {10.1038/sj.bjc.6601118}
}

@article{vandenpoel2004customerattritionanalysis,
title = {{Customer attrition analysis for financial services using proportional hazard models}},
journal = {European Journal of Operational Research},
volume = {157},
number = {1},
pages = {196-217},
year = {2004},
issn = {0377-2217},
doi = {10.1016/S0377-2217(03)00069-9},
author = {Dirk {Van den Poel} and Bart Larivière},
}

@book{meeker1998statisticalmethods,
  title     = {Statistical Methods for Reliability Data},
  author    = {Meeker, William Q. and Escobar, Luis A.},
  year      = {1998},
  publisher = {John Wiley \& Sons},
}

@article{qi2026survivalpfnamortizingsurvivalprediction,
      title={{SurvivalPFN: Amortizing Survival Prediction via In-Context Bayesian Inference}}, 
      author={{Shi-ang} Qi and Vahid Balazadeh and Michael Cooper and Russell Greiner and Rahul G. Krishnan},
    journal={arXiv:2605.15488},
    year={2026}
}

@article{seletkov2026survivalincontextamortizedbayesian,
  title={{Survival In-Context: Amortized Bayesian Survival Analysis via Prior-Fitted Networks}},
      author={Dmitrii Seletkov and Paul Hager and Georgios Kaissis and Rickmer Braren and Daniel Rueckert and Raphael Rehms},
  journal={arXiv:2603.29475},
  year={2026}
}

@article{kim2026tabularfoundationmodelssurvival,
  title={Tabular foundation models can do survival analysis},
  author={Kim, Da In and Lai, Wei Siang and Zhang, Kelly W},
  journal={arXiv:2601.22259},
  year={2026}
}

@ARTICLE{glymour2019reviewofcausaldiscoverymethods,
AUTHOR={Glymour, Clark  and Zhang, Kun  and Spirtes, Peter},
TITLE={{Review of Causal Discovery Methods Based on Graphical Models}},
JOURNAL={Frontiers in Genetics},
VOLUME={Volume 10},
YEAR={2019},
DOI={10.3389/fgene.2019.00524},
ISSN={1664-8021},
}

@article{kalish2007estimatinghighdimensional,
  author  = {Markus Kalisch and Peter B{{\"u}}hlmann},
  title   = {{Estimating High-Dimensional Directed Acyclic Graphs with the PC-Algorithm}},
  journal = {Journal of Machine Learning Research},
  year    = {2007},
  volume  = {8},
  number  = {22},
  pages   = {613--636},
}

@article{qiao2026cdfmgeneralpurposecausaldiscovery,
      title={{CDFM: Towards a General-Purpose Causal Discovery Foundation Model}}, 
      author={Jie Qiao and Ruichu Cai and Zijian Li and Weilin Chen and Pengfei Hua and Boyan Xu and Zhengming Chen and Zhifeng Hao and Peng Cui},
      year={2026},
      journal={arXiv:2607.11508},
}

@article{thompson2026arrowfoundationmodelcausal,
      title={{Arrow: A Foundation Model for Causal Discovery}}, 
      author={Ryan Thompson and He Zhao and Daniel M. Steinberg and Edwin V. Bonilla},
      year={2026},
      journal={arXiv:2605.07204},
}

@article{li2026tabcausalpretrainingcausalenvironments,
      title={{TabCausal: Pretraining Across Causal Environments for Tabular Causal Discovery}}, 
      author={Zi-Rong Li and Si-Yang Liu and Tian-Zuo Wang and Han-Jia Ye},
      year={2026},
      journal={arXiv:2605.31156},
}

@article{
wu2025sampleestimateaggregaterecipe,
title={{Sample, estimate, aggregate: A recipe for causal discovery foundation models}},
author={Menghua Wu and Yujia Bao and Regina Barzilay and Tommi Jaakkola},
journal={Transactions on Machine Learning Research},
issn={2835-8856},
year={2025},
}

@article{guan2026dcdpfndecouplingawarefoundationmodel,
      title={{DCD-PFN: A Decoupling-Aware Foundation Model for Causal Discovery}}, 
      author={Zhengkang Guan and Yikang Chen and Yi He and Yunze Tong and Zijing Hu and Haoyuan Qian and Fei Wu and Kun Kuang},
      year={2026},
      journal={arXiv:2606.21212},
}

@article{chen2026dagfmfoundationmodelcausal,
      title={{DAG-FM: A Foundation Model for Causal Discovery under Heterogeneous Causal Mechanisms}}, 
      author={Yikang Chen and Zhengkang Guan and Haoyuan Qian and Xingxuan Zhang and Peng Cui and Yi Yang and Fei Wu and Kun Kuang},
      year={2026},
      journal={arXiv:2607.11510}, 
}

@article{blobaum2026foundcausecausaldiscoverylatent,
      title={{FoundCause: Causal Discovery with Latent Confounders from Observational Data}}, 
      author={Patrick Blöbaum and Krishnakumar Balasubramanian and Shiva Prasad Kasiviswanathan},
      year={2026},
      journal={arXiv:2606.17516},
}

@article{bohm2026survpfnfoundationmodelssurvival,
      title={{SurvPFN: Towards Foundation Models for Survival Predictions}}, 
      author={Samuel Böhm and Lennart Purucker and Frank Hutter and Pascal Schlosser},
      year={2026},
      journal={arXiv:2606.04564}, 
}

@InProceedings{reuter2026use,
title={{Use What You Know: Causal Foundation Models with Partial Graphs}},
author={Arik Reuter and Anish Dhir and Cristiana Diaconu and Jake Robertson and Ole Ossen and Frank Hutter and Adrian Weller and Mark van der Wilk and Bernhard Sch{\"o}lkopf},
  booktitle = {Proceedings of the 43rd International Conference on Machine Learning},
  year = 	 {2026},
  series = 	 {Proceedings of Machine Learning Research},
}

@article{thumm2026continuoustimecausalfoundationmodels,
      title={{Towards Continuous-time Causal Foundation Models}}, 
      author={Dennis Thumm and Ruben Wiedemann and Ying Chen},
      year={2026},
      journal={arXiv:2605.28880}, 
}

@article{thumm2026interventionaltimeseries,
  title={{Interventional Time Series Priors for Causal Foundation Models}},
  author={Thumm, Dennis and Chen, Ying},
  journal={arXiv:2603.11090},
  year={2026}
}

@article{bellot2026foundationmodelspartialcausal,
      title={{Foundation Models for Partial Causal Identification}}, 
      author={Alexis Bellot and Anish Dhir},
      year={2026},
      journal={arXiv:2608.20841},
}

@inproceedings{dhir2025estimatinginterventionaldistributions,
 author = {Dhir, Anish and Diaconu, Cristiana and Lungu, Valentinian and Requeima, James and Turner, Richard and van der Wilk, Mark},
 booktitle = {Advances in Neural Information Processing Systems},
 doi = {10.52202/085713-4681},
 pages = {140060--140096},
 title = {{Estimating Interventional Distributions with Uncertain Causal Graphs through Meta-Learning}},
 volume = {38},
 year = {2025}
}

@inproceedings{kougioulis2026largecausalmodelstemporal,
author="Nikolaos Kougioulis and Nikolaos Gkorgkolis and MingXue Wang and Bora Caglayan and Dario Simionato and Andrea Tonon and Ioannis Tsamardinos",
title="{{Large Causal Models for Temporal Causal Discovery}}",
  booktitle={Joint European Conference on Machine Learning and Knowledge Discovery in Databases},
year="2026",
  organization={Springer}
}

@article{
montagna2026demystifyingamortizedcausaldiscovery,
title={{Demystifying amortized causal discovery with transformers}},
author={Francesco Montagna and Max Cairney-Leeming and Dhanya Sridhar and Francesco Locatello},
journal={Transactions on Machine Learning Research},
issn={2835-8856},
year={2025},
}

@InProceedings{melnychuk2026frequentistconsistencypriordatafitted,
      title={{Frequentist Consistency of Prior-Data Fitted Networks for Causal Inference}}, 
      author={Valentyn Melnychuk and Vahid Balazadeh and Stefan Feuerriegel and Rahul G. Krishnan},
  booktitle = 	 {Proceedings of the 43nd International Conference on Machine Learning},
  year = 	 {2026},
  series = 	 {Proceedings of Machine Learning Research},
}

@inproceedings{
mahajan2024empirical,
title={{Empirical Analysis of Model Selection for Heterogeneous Causal Effect Estimation}},
author={Divyat Mahajan and Ioannis Mitliagkas and Brady Neal and Vasilis Syrgkanis},
booktitle={The Twelfth International Conference on Learning Representations},
year={2024},
}

@inproceedings{spinaci2025contexttab,
 author = {Spinaci, Marco and Polewczyk, Marek and Schambach, Maximilian and Thelin, Sam},
 booktitle = {Advances in Neural Information Processing Systems},
 doi = {10.52202/085713-4918},
 title = {{ConTextTab: A Semantics-Aware Tabular In-Context Learner}},
 volume = {38},
 year = {2025}
}

@article{garg2025real,
      title={{Real-TabPFN: Improving Tabular Foundation Models via Continued Pre-training With Real-World Data}}, 
      author={Anurag Garg and Muhammad Ali and Noah Hollmann and Lennart Purucker and Samuel Müller and Frank Hutter},
  journal={arXiv:2507.03971},
  year={2025}
}

@inproceedings{
manela2024marginal,
title={{Marginal Causal Flows for Validation and Inference}},
author={Daniel de Vassimon Manela and Laura Battaglia and Robin J. Evans},
booktitle={The Thirty-eighth Annual Conference on Neural Information Processing Systems},
year={2024},
}

@article{mourao2026priordatafittednetworkscausal,
      title={{Prior-Data Fitted Networks for Causal Inference: a Simulation Study with Real-World Scenarios}}, 
      author={Francisco Mourao and David Hajage and Daria Bystrova and Bertrand Bouvarel and Nathanaël Lapidus and Fabrice Carrat and Benjamin Glemain},
      year={2026},
      journal={arXiv:2603.15928},
}

@inproceedings{
wang2026unveilingpriordatafittednetworks,
title={Unveiling Prior-Data Fitted Networks on Causal Effect Estimation: Pre-Training or Fine-Tuning?},
author={Haotian Wang and Xinpeng Lv and Hao Zou and Yanghao Xiao and Shanzhi Gu and Yang Shi and Yunxin Mao and Yuanxing Zhang and Mingyang Geng and Shaowu Yang and Haoxuan Li and Wenjing Yang and Peng Cui and Zhouchen Lin},
booktitle={Forty-third International Conference on Machine Learning},
year={2026},
}

@article{vanderlaan2026doublyrobustinferencecalibration,
      title={Doubly robust inference via calibration}, 
      author={Lars van der Laan and Alex Luedtke and Marco Carone},
      year={2026},
      journal={arXiv:2411.02771},
}

@article{gao2026perturbpfnprobinglimitssynthetic,
      title={{PerturbPFN: Probing the Limits of Synthetic Priors in Drug Perturbation Modelling}}, 
      author={Yuche Gao and José Miguel Hernández-Lobato and Siyuan Guo},
      year={2026},
      journal={arXiv:2607.23447},
}

@article{yin2025learningcausalgraphsscale,
      title={{Learning Causal Graphs at Scale: A Foundation Model Approach}}, 
      author={Naiyu Yin and Tian Gao and Yue Yu},
      year={2025},
      journal={arXiv:2506.18285},
}

@book{rosenbaum2002observational,
  title     = {Observational Studies},
  author    = {Rosenbaum, Paul R.},
  edition   = {2},
  year      = {2002},
  publisher = {Springer},
  address   = {New York},
  doi       = {10.1007/978-1-4757-3692-2}
}

@inproceedings{
ham2026astructuralviewofquery,
title={A Structural View of Query Misspecification in Causal Foundation Models},
author={Junha Ham and Deokgyu Kim and Doeun Kim and Serjin Kim and Sanghack Lee},
booktitle={ICML 2026 Workshop on Structured Probabilistic Inference {\&} Generative Modeling},
year={2026},
}

@inproceedings{
thumm2026causalfoundationmodelsfortimeseries,
title={Causal Foundation Models for Time Series based on Prior-Data fitted Networks},
author={Dennis Thumm and Arik Reuter and Jake Robertson and Shi Bin Hoo and Adrian Weller and Frank Hutter and Ying Chen and Bernhard Sch{\"o}lkopf},
booktitle={2nd ICML Workshop on Foundation Models for Structured Data},
year={2026},
}

@article{bunne2023learningsinglecell,
author = {Bunne, Charlotte and Stark, Stefan G. and Gut, Gabriele and del Castillo, Jacobo Sarabia and Levesque, Mitch and Lehmann, Kjong-Van and Pelkmans, Lucas and Krause, Andreas and Rätsch, Gunnar},
title = {Learning single-cell perturbation responses using neural optimal transport},
journal = {Nature Methods},
year = {2023},
volume = {20},
number = {11},
doi = {10.1038/s41592-023-01969-x},
issn = {1548-7105}
}

@article{wang2021flamlfastlightweightautoml,
      title={FLAML: A Fast and Lightweight AutoML Library}, 
      author={Chi Wang and Qingyun Wu and Markus Weimer and Erkang Zhu},
      year={2021},
      journal={arXiv:1911.04706},
}

@article{smith2005does,
  title   = {Does matching overcome {LaLonde}'s critique of nonexperimental estimators?},
  author  = {Smith, Jeffrey A. and Todd, Petra E.},
  journal = {Journal of Econometrics},
  volume  = {125},
  number  = {1--2},
  pages   = {305--353},
  year    = {2005},
  issn    = {0304-4076},
  doi     = {10.1016/j.jeconom.2004.04.011},
  publisher = {Elsevier}
}

\end{document}